\documentclass[11pt]{article}

\usepackage[final]{acl}

\usepackage{times}
\usepackage{latexsym}

\usepackage[T1]{fontenc}

\usepackage[utf8]{inputenc}

\usepackage{microtype}

\usepackage{inconsolata}

\usepackage{graphicx}
\usepackage{subcaption}
\usepackage{enumitem}
\usepackage{amsmath}
\usepackage{amssymb}
\usepackage{tcolorbox}
\usepackage{booktabs}
\usepackage{multirow}
\usepackage[table]{xcolor}
\newcommand{\rowhcolor}{\rowcolor[rgb]{0.929, 0.961, 0.980}}
\newcommand{\gain}[1]{\textcolor[RGB]{210, 50, 45}{$_{\uparrow #1}$}}
\newcommand{\hgain}[1]{\textcolor[RGB]{210, 50, 45}{$_{\mathbf{\uparrow #1}}$}}
\newcommand{\drop}[1]{\textcolor[rgb]{0.18,0.62,0.22}{$_{\downarrow #1}$}}

\title{ReverseEOL: Improving Training-free Text Embeddings via\\ Text Reversal in Decoder-only LLMs}

\author{
    Ailiang Lin$^{1}$, 
    Zhuoyun Li$^{2}$,
    Yusong Wang$^{1}$,
    Keyu Mao$^{1}$,\\
    \textbf{Kotaro Funakoshi}$^{1}$,
    \textbf{Manabu Okumura}$^{1}$\\
    $^{1}$Institute of Science Tokyo \quad $^{2}$Tencent\\
\texttt{\{linailiang, wangyi, maokeyu, funakoshi, oku\}@lr.first.iir.isct.ac.jp}\\ \texttt{earyli@tencent.com}
}

\begin{document}
\maketitle
\begin{abstract}
Recent advances in Large Language Models (LLMs) have opened new avenues for generating training-free text embeddings. However, the causal attention in decoder-only LLMs prevents earlier tokens from attending to future context, leading to biased contextualized representations. In this work, we propose \textit{Reverse prompting with Explicit One-word Limitation} (ReverseEOL), a simple yet effective method for enhancing the representational capability of frozen LLMs. ReverseEOL augments the standard forward embedding with an additional reversed embedding derived from the reversed input text. Since reversing the input exposes each token to context inaccessible in the original order, the resulting reversed embedding effectively provides complementary information to the original one. As a result, combining the forward and reversed embeddings yields a richer final representation. Comprehensive experiments on STS and MTEB benchmarks demonstrate that ReverseEOL significantly improves the performance of existing training-free baselines across a broad range of LLMs with diverse architectures and scales. Extensive ablations and analyses further confirm the necessity of our reversal mechanism.
\end{abstract}

\section{Introduction}
Text embeddings~\cite{embeddingsurvey} are foundational to a wide range of natural language processing (NLP) tasks, including question answering, semantic textual similarity, and information retrieval~\cite{mteb}. While Large Language Models (LLMs) have demonstrated remarkable performance across various embedding benchmarks, adapting them into dedicated embedding models typically requires extensive training data and sophisticated engineering optimizations~\cite{geminiembedding,qwen3embedding}. In contrast, eliciting text embeddings from frozen LLMs offers a practical and compelling alternative, as it requires no parameter updates and can be directly applied to off-the-shelf LLMs.

Training-free text embeddings are typically derived from the last token in the sequence. However, this token is primarily optimized for next-token prediction rather than for capturing the semantics of the input. To bridge the mismatch between the role of the last token in generation and embedding tasks, recent prompt-based studies~\cite{metaeol,prompteol,cot} explicitly instruct LLMs with the one-word limitation. A representative prompt is \textit{This sentence : ``[TEXT]'' means in one word:}", which encourages LLMs to condense the semantic information into a single token.

However, most LLMs adopt a decoder-only architecture with causal attention, which prevents earlier tokens from accessing future context. As a result, the last token aggregates information from preceding tokens that encode only incomplete semantic information, resulting in biased contextualized representations. To mitigate this issue, Contrastive Prompting~\cite{cp} filters out non-essential information by directly modifying the last token, while some studies~\cite{tp, kvembedding} prepend context-aware tokens to the sequence. Nevertheless, these techniques typically require internal model interventions at carefully selected layers, limiting their applicability across diverse LLM backbones. More importantly, they achieve only marginal gains over the corresponding prompt-based baselines.

To this end, we propose \underline{\textbf{Reverse}} prompting with \underline{\textbf{E}}xplicit \underline{\textbf{O}}ne-word \underline{\textbf{L}}imitation (\textbf{ReverseEOL}), a simple yet effective approach for enhancing training-free text embeddings. Our method stems from a straightforward intuition: if each token can capture only limited context under causal attention, reversing the input allows the same token to access complementary information from the opposite direction. Specifically, we first obtain a forward text embedding based on the given prompt in the standard manner. We then reverse the input text and adapt the prompt accordingly to derive an additional reversed embedding, which complements the forward embedding by capturing contextual information unavailable in the original order. Finally, we average the two embeddings to generate a higher-quality representation. ReverseEOL requires no complex prompt engineering and is model-agnostic, allowing it to be seamlessly integrated with a wide range of training-free text embedding methods.

We conduct comprehensive evaluations on STS and MTEB~\cite{mteb} benchmarks across more than 10 LLM families with parameter sizes ranging from 0.5B to 32B, covering both dense and Mixture-of-Experts (MoE) architectures. Experimental results show that ReverseEOL significantly improves existing training-free baselines by up to $8.09$ and $5.34$ points on the two benchmarks, respectively. Extensive ablations and analyses further confirm that the proposed reversed embedding effectively complements the forward one with more diverse semantic information, thereby consistently improving the overall representation quality.

Our main contributions are as follows:
\begin{itemize}
\item We propose ReverseEOL, a simple yet effective method that enhances training-free text embeddings without complex prompt engineering or internal model interventions.
\item ReverseEOL combines the forward text embedding with a complementary embedding derived from the reversed input text, thereby incorporating more diverse contextual information and improving representation quality.
\item Extensive experiments on STS and MTEB benchmarks across more than 10 LLM families demonstrate that ReverseEOL significantly improves existing training-free baselines. We further present in-depth ablations and analyses to reveal the mechanism behind the effectiveness of our method.
\end{itemize}

\section{Related Work}
\paragraph{LLM-based Text Embeddings.}
Text embeddings play a critical role in various NLP applications, such as semantic search and retrieval-augmented
generation (RAG). Early efforts on text embeddings primarily relied on encoder-only language models like BERT~\cite{bert} and RoBERTa~\cite{roberta}. With recent advances in LLMs, considerable efforts have shifted to adapting them into dedicated text embedding models. A line of work~\cite{llm2vec,anchor,causal2vec} fine-tunes LLMs with contrastive learning on publicly available retrieval datasets, while methods such as E5-Mistral~\cite{e5mistral} and MGH~\cite{mgh} further leverage LLM-generated synthetic training data to improve embedding performance. Beyond academic efforts, many industry-developed embedding models, including KaLM-Embedding~\cite{kalm}, Gemini Embedding~\cite{geminiembedding}, and Qwen3 Embedding~\cite{qwen3embedding}, achieve state-of-the-art performance through large-scale curated and synthetic data together with sophisticated engineering optimizations. Despite their remarkable performance on embedding benchmarks, fine-tuning LLMs inevitably incurs non-trivial computational costs and may lead to catastrophic forgetting of their original general capabilities. 

\paragraph{Training-free Text Embeddings.}
An appealing and practical alternative is to elicit text embeddings from frozen LLMs in a training-free manner. PromptEOL~\cite{prompteol} pioneers this direction by introducing the one-word limitation prompt, which encourages LLMs to condense sentence semantics into a single token. Subsequent efforts further enhance the prompt design from various perspectives: Pretended CoT~\cite{cot} prepends a chain-of-thought instruction, Knowledge~\cite{cot} incorporates summarization-related prior knowledge, and MetaEOL~\cite{metaeol} aggregates representations from multiple meta-task prompts. In addition, ECHO~\cite{echo} duplicates the input and extracts contextualized representations from the second occurrence. Beyond prompt engineering, some approaches modify the internal computation of LLMs to alleviate the limitation of causal attention. Contrastive Prompting~\cite{cp} steers intermediate activations using an auxiliary prompt at inference time, while methods such as Token Prepending~\cite{tp} and KV-Embedding~\cite{kvembedding} prepend a context-aware token to the internal states. In contrast, our ReverseEOL augments the original text embedding with complementary information from the reversed input, requiring no elaborate prompt design or internal model interventions, making it readily applicable to various existing training-free baselines across diverse LLMs.

\begin{figure*}[htbp]
\centering
\includegraphics[width=0.99\textwidth]{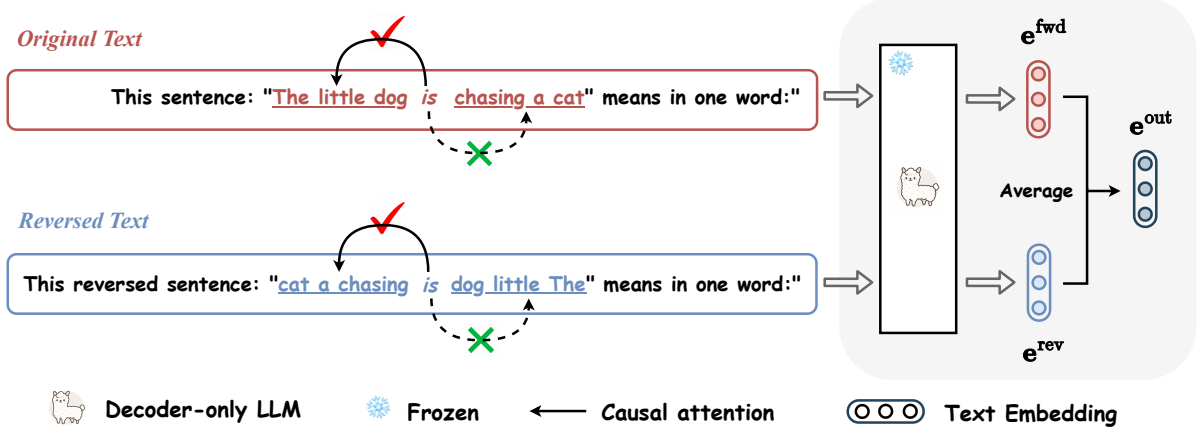}
\caption{Overview of ReverseEOL. Given the input  \textit{``The little dog is chasing a cat''}, we feed it into a frozen LLM to obtain the forward embedding $\mathbf{e}^{\mathrm{fwd}}$, and process its reversed version \textit{``cat a chasing is dog little The''} to derive an additional reversed embedding $\mathbf{e}^{\mathrm{rev}}$. In this way, the same word (e.g., \textit{is}) sees complementary contexts under causal attention in the two orders. The final embedding $\mathbf{e}^{\mathrm{out}}$ is obtained by averaging $\mathbf{e}^{\mathrm{fwd}}$ and $\mathbf{e}^{\mathrm{rev}}$.}
\label{fig:model}
\end{figure*}

\section{Method}
\subsection{Preliminary}
LLMs have been widely adopted for generating text embeddings due to their remarkable semantic understanding capabilities. A decoder-only LLM typically consists of a tokenizer and an embedding layer, which we jointly denote as $\mathcal{E}(\cdot)$, followed by $L$ sequential decoder blocks $\mathrm{Dec}(\cdot)$. Given a raw input text $x$, we first map it into a sequence of token representations $\mathcal{E}(x)= (\mathbf{x}_1, \mathbf{x}_2, \ldots, \mathbf{x}_n)\in \mathbb{R}^{n\times d}$ of length $n$ with embedding dimension $d$, which are then passed into the decoder layers. Training-free text embeddings are typically obtained via last-token pooling $\mathrm{Pool}_{\textsc{last}}$ at a selected layer $\ell \in \{1, \ldots, L\}$, which takes the hidden state of the last token as the final text representation, formally defined as:
\begin{equation}
\mathbf{e}_{x} = \mathrm{Pool}_{\textsc{last}}(f_\theta(x)) \in \mathbb{R}^d, \\
\end{equation}
\begin{equation}
f_\theta(\cdot) = \mathrm{Dec}_\ell(\mathrm{Dec}_{\ell-1}(\ldots \mathrm{Dec}_1(\mathcal{E}(\cdot)))).
\end{equation}

Notably, the last-token representation $\mathbf{e}_{x}$ in a frozen LLM is inherently biased towards predicting the next token rather than capturing the semantic information of the input. To mitigate this misalignment, recent studies explicitly prompt LLMs to condense the meaning of the input into a single word. For example, PromptEOL~\cite{prompteol} employs the following prompt:
\begin{tcolorbox}[colback=blue!3, colframe=blue!15, boxrule=0.5pt, arc=4pt, left=8pt, right=8pt, top=8pt, bottom=8pt]
\fontsize{9.4pt}{12pt}\selectfont
\texttt{This sentence : ``[Text]'' means in one word:}"
\end{tcolorbox}
\noindent where \texttt{[Text]} denotes the placeholder for the input text. The one-word limitation effectively leverages the generative capability of LLMs to compress the contextualized information, thereby bridging the gap between the roles of the last token in generation and embedding tasks. By wrapping the input text in a prompt template $P(\cdot)$, the forward text embedding is obtained as:
\begin{equation}
\mathbf{e}_{x}^{\text{fwd}} = \mathrm{Pool}_{\textsc{last}}(f_\theta(P(\mathbf{x}))). 
\end{equation}

\subsection{ReverseEOL}
While the one-word limitation prompt effectively guides the last token toward semantic summarization, it fails to address the fundamental bottleneck of causal attention in representation learning: tokens preceding the last position encode only partial context, causing incomplete semantic information to propagate into the final representation~\cite{tp}. To mitigate this limitation, we propose ReverseEOL, a simple yet effective method that enhances training-free text embeddings by augmenting the original forward embedding $\mathbf{e}_{x}^{\text{fwd}}$ with a complementary embedding derived from the reversed input text.

Our method stems from a simple intuition: if each token can only access limited context in the original order, reversing the input allows the same token to access complementary information from the opposite direction. Specifically, as shown in Figure~\ref{fig:model}, we perform reversal on the original input text $x$ to obtain the corresponding reversed text $x^{\text{rev}}$, and adapt the original prompt template $P(\cdot)$ into a reversed variant $P^{\text{rev}}(\cdot)$ that explicitly instructs the LLM to process the reversed input. Taking PromptEOL as an example, the prompt can be modified as follows:
\begin{tcolorbox}[colback=blue!3, colframe=blue!15, boxrule=0.5pt, arc=4pt, left=8pt, right=8pt, top=8pt, bottom=8pt]
\fontsize{9.5pt}{12pt}\selectfont
\texttt{This reversed sentence : ``[Reversed\_Text]'' means in one word:}"
\end{tcolorbox}
\noindent where \texttt{[Reversed\_Text]} is the placeholder for $x^{\text{rev}}$. The reversed embedding is computed as:
\begin{equation}
\mathbf{e}_{x}^{\text{rev}} = \mathrm{Pool}_{\textsc{last}}(f_\theta(P^{\text{rev}}(x^{\text{rev}}))). 
\end{equation}

\begin{figure}[t]
    \centering
    \includegraphics[width=\columnwidth]{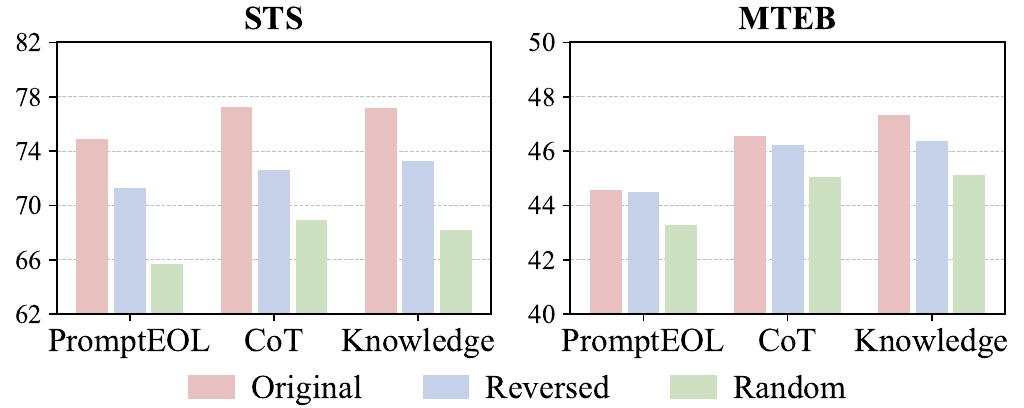}
    \caption{Performance comparison of original, reversed, and randomly shuffled inputs on STS and MTEB across three training-free baselines using LLaMA-2-7B. CoT refers to Pretended CoT.}
    \label{fig:ori_reverse_random}
\end{figure}
This raises a natural question: \textit{can LLMs truly understand reversed text?} As shown in Figure~\ref{fig:ori_reverse_random}, the reversed embedding $\mathbf{e}_{x}^{\text{rev}}$ alone yields lower performance than the forward one $\mathbf{e}_{x}^{\text{fwd}}$, since reversing inevitably disrupts the original sequential structure. Nevertheless, it still substantially outperforms embeddings derived from randomly shuffled input, confirming that LLMs can extract meaningful semantics from reversed text.

Finally, we average the two embeddings obtained from the original and reversed orders to produce the final representation:
\begin{equation}
\mathbf{e}_{x}^{\text{out}}= \frac{1}{2}(\mathbf{e}_{x}^{\text{fwd}} + \mathbf{e}_{x}^{\text{rev}}). 
\end{equation}
Notably, ReverseEOL requires neither complex prompt engineering nor internal model interventions. It simply adapts an existing one-word limitation prompt (e.g., PromptEOL) by indicating that the input text is reversed, making it readily applicable to a wide range of existing training-free baselines across diverse LLMs. Furthermore, we present detailed analyses in Section~\ref{sec:mechanism_analysis} to demonstrate that the reversed embedding $\mathbf{e}_{x}^{\text{rev}}$ effectively complements the forward embedding $\mathbf{e}_{x}^{\text{fwd}}$ with richer semantic information, thereby significantly improving text embedding performance.

\begin{table*}[th]
\centering
\scalebox{0.86}{
\small
\setlength{\tabcolsep}{12pt}
\begin{tabular}{lcccccccc}
\toprule
\textbf{Method} & \textbf{STS12} & \textbf{STS13} & \textbf{STS14} & \textbf{STS15} & \textbf{STS16} & \textbf{STS-B} & \textbf{SICK-R} & \textbf{Avg.}\\
\midrule
LLaMA-2-7B$_{\text{last-token}}$ &14.85 &32.89 &18.53 &20.63 &52.65 &31.54 &52.09 &31.88  \\
LLaMA-2-7B$_{\text{mean}}$ &35.49 &53.15 &40.12 &55.35 &53.26 &42.10 &49.96 &47.06  \\
CP$^{\dag}$ &67.79 &83.66 &74.52 &81.10 &80.70 &80.39 &74.01 &77.45 \\
TP$^{\dag}$ &68.52 &83.44 &75.23 &79.36 &81.33 &80.37 &74.51 &77.54 \\
MetaEOL &69.93 &82.85 &75.75 &83.90 &80.82 &81.67 &70.70 &77.95 \\
\midrule
PromptEOL &64.36 &82.05 &72.21 &78.88 &77.35 &76.80 &72.52 &74.88 \\
\rowhcolor\quad \textit{w/ ReverseEOL} &70.71 &84.57 &76.14 &83.39 &80.65 &81.72 &73.64 &78.69 \textcolor[rgb]{0.86,0.20,0.18}{\textbf{(+3.81)}}\\
\midrule
ECHO  &67.38 &81.08 &71.20 &78.92 &77.61 &77.82 &72.21 &75.17 \\
\rowhcolor\quad \textit{w/ ReverseEOL} &\textbf{72.24} &84.69 &75.15 &83.29 &81.10 &81.98 &74.93 &79.05 \textcolor[rgb]{0.86,0.20,0.18}{\textbf{(+3.88)}}\\
\midrule
Pretended CoT &67.76 &83.87 &74.73 &79.94 &80.85 &79.86 &73.59 &77.23 \\
\rowhcolor\quad \textit{w/ ReverseEOL} &72.20 &85.51 &\textbf{77.51} &\textbf{83.92} &82.10 &\textbf{82.80} &74.50 &79.79 \textcolor[rgb]{0.86,0.20,0.18}{\textbf{(+2.56)}}\\
\midrule
Knowledge &65.60 &82.82 &74.48 &80.75 &80.13 &80.34 &75.89 &77.14 \\
\rowhcolor\quad \textit{w/ ReverseEOL} &71.50 &\textbf{85.56} &77.39 &83.74 &\textbf{82.30} &82.51 &\textbf{75.98} &\textbf{79.85} \textcolor[rgb]{0.86,0.20,0.18}{\textbf{(+2.71)}}\\
\bottomrule
\end{tabular}}
\caption{Results on STS tasks using LLaMA-2-7B. \dag~denotes using Pretended CoT as the base prompt. The best results are highlighted in \textbf{bold}. Results with other LLM backbones are provided in Table~\ref{tab:prompt_roubust} (Appendix~\ref{appendix:prompt_robust}).}
\label{tab:sts_compare}
\end{table*}

\section{Experiments}
\subsection{Experimental Setup}
\paragraph{Benchmarks.} We evaluate ReverseEOL on two widely used benchmarks. The first is Semantic Textual Similarity (STS), the most commonly adopted benchmark for training-free text embedding methods, which consists of seven datasets, including STS 2012--2016~\cite{sts12,sts13,sts14,sts15,sts16}, STS-B~\cite{stsb}, and SICK-R~\cite{sickr}. To further assess the generalizability of training-free methods, we additionally evaluate it across a broader range of downstream tasks from the Massive Text Embedding Benchmark (MTEB)~\cite{mteb}, comprising five task categories: Classification (7 datasets), Pair Classification (3 datasets), Reranking (3 datasets), Clustering (10 datasets), and Retrieval (9 datasets). More details about these benchmarks are provided in Appendices~\ref{appendix:mteb_detail} and~\ref{appendix:sts_detail}.

\paragraph{Implementation.} We conduct experiments and ablation studies across more than 10 decoder-only LLM families with parameter sizes ranging from 0.5B to 32B, covering both dense and MoE architectures. More details about these LLMs can be found in Appendix~\ref{appendix:llm_detail}. We adopt an early-exit strategy that extracts embeddings from an intermediate layer rather than the final layer. Specifically, we use the 27th layer ($\ell=27$) for LLaMA-2-7B and the penultimate layer ($\ell=L-1$) for all other LLMs, following prior work~\cite{cot,tp,cp}. To ensure a fair comparison, all competing methods within the same experiment share the identical early-exit setting and evaluation pipeline. A detailed analysis of the exit layer is provided in Appendix~\ref{appendix:exit_layer}. The maximum input sequence length is set to 512 tokens, and we use a single NVIDIA A100 80GB GPU for all experiments.

\paragraph{Baselines.} The training-free text embedding baselines fall into two categories. (i) \textit{Prompt Engineering:} \textbf{PromptEOL}~\cite{prompteol} is the first to introduce the one-word limitation prompt to condense the input semantics into a single token. \textbf{MetaEOL}~\cite{metaeol} aggregates representations produced by eight task-specific prompts. \textbf{ECHO}~\cite{echo} duplicates the input and extracts embeddings from the repeated tokens. \textbf{Pretended CoT}~\cite{cot} extends PromptEOL with zero-shot chain-of-thought prompting. \textbf{Knowledge}~\cite{cot} incorporates human prior knowledge into the prompt. For most prompt-based methods, ReverseEOL can be directly integrated by constructing their reversed variant, with more details provided in Appendix~\ref{appendix:forward_reversed_prompt}. (ii) \textit{Internal Model Intervention:} Built upon prompt engineering methods, \textbf{Contrastive Prompting (CP)}~\cite{cp} steers text embeddings toward core semantics by directly modifying the last token. \textbf{Token Prepending (TP)}~\cite{tp} prepends the previous layer’s last token to the next layer's input sequence.

\subsection{Results}
\paragraph{STS Results.} Table~\ref{tab:sts_compare} presents the results on STS tasks. Integrating ReverseEOL into four prompt-based methods consistently improves performance across all seven datasets, with average gains ranging from $2.56$ to $3.88$ points. In particular, ReverseEOL boosts the lowest-performing baseline PromptEOL from $74.88$ to $78.69$, surpassing all previous methods. When built upon Pretended CoT, CP and TP achieve only marginal gains of $0.22$ and $0.31$ points, whereas ReverseEOL delivers a substantial improvement of $2.56$ points under the same setting. Moreover, our best-performing variant, Knowledge \textit{w/ ReverseEOL}, reaches an average score of $79.85$, outperforming the prior strongest baseline MetaEOL ($77.95$) by a clear margin. Notably, MetaEOL aggregates eight embeddings produced by carefully designed prompts, whereas ReverseEOL requires only an additional reversed embedding, highlighting the simplicity and effectiveness of our reversal strategy.
\paragraph{MTEB Results.} Beyond STS tasks, we further assess the generalizability of ReverseEOL across five embedding task categories on MTEB. As shown in Table~\ref{tab:mteb_compare}, ReverseEOL, applied to PromptEOL, consistently outperforms all competing baselines, yielding average improvements of $5.34$, $5.03$, and $4.18$ points on LLaMA-2-7B, LLaMA-2-13B, and Mistral-7B, respectively. The gains are even more pronounced on retrieval tasks that involve long input texts, where our method achieves improvements of $7.23$, $6.29$, and $5.23$ points. This suggests that ReverseEOL effectively mitigates the bottleneck of causal attention in long-sequence scenarios. Notably, under the same setting, both TP and CP even degrade the performance of PromptEOL on Mistral-7B, indicating that existing training-free methods may struggle to produce versatile embeddings. In contrast, ReverseEOL remains consistently effective across diverse embedding tasks without requiring complex prompt engineering or internal model interventions. See Appendix~\ref{appendix:mteb_results} for detailed results for each dataset.

\begin{table*}[th]
\centering
\begin{minipage}[t]{0.55\textwidth}
\centering
\scalebox{0.79}{%
\small
\setlength{\tabcolsep}{6pt}
\begin{tabular}{lcccccc}
\toprule
\textbf{Method} & \textbf{Retr.} & \textbf{Rerank.} & \textbf{Clust.} & \textbf{PairClass.} & \textbf{Class.} & \textbf{Avg.}\\
\midrule
\multicolumn{7}{c}{\textbf{\texttt{LLaMA-2-7B}}}\\\midrule
ECHO &25.73 &59.15 &38.06 &\textbf{74.04} &71.60 &47.28 \\
PromptEOL &26.92 &57.94 &31.94 &62.36 &71.97 &44.57\\
\quad \textit{w/ CP}  &28.22 &58.41 &33.20 &63.31 &72.45 &45.57\\
\quad \textit{w/ TP}  &29.84 &58.81 &34.14 &64.65 &72.80 &46.56\\
\rowhcolor\quad \textit{w/ ReverseEOL} &\textbf{34.15} &\textbf{59.75} &\textbf{38.64} &69.67 &\textbf{73.61} &\textbf{49.91}\hgain{5.34}\\
\midrule
\multicolumn{7}{c}{\textbf{\texttt{LLaMA-2-13B}}}\\\midrule
ECHO &24.77 &53.08 &31.98 &\textbf{72.03} &68.70 &43.72 \\
PromptEOL &25.66 &58.59 &31.77 &57.63 &70.87 &43.54\\
\quad \textit{w/ CP}  &25.11 &58.70 &32.11 &57.21 &70.47 &43.38\\
\quad \textit{w/ TP}  &26.28 &59.67 &33.17 &60.82 &71.83 &44.76\\
\rowhcolor\quad \textit{w/ ReverseEOL} &\textbf{31.95} &\textbf{60.06} &\textbf{36.95} &69.03 &\textbf{72.81} &\textbf{48.57}\hgain{5.03}\\
\midrule
\multicolumn{7}{c}{\textbf{\texttt{Mistral-7B}}}\\\midrule
ECHO &21.41 &52.02 &34.29 &70.55 &68.59 &43.23 \\
PromptEOL &28.06 &60.39 &36.21 &61.09 &71.55 &46.25\\
\quad \textit{w/ CP}  &27.14 &60.54 &35.68 &60.78 &71.43 &45.79\\
\quad \textit{w/ TP}  &28.21 &60.18 &35.69 &61.93 &71.71 &46.22\\
\rowhcolor\quad \textit{w/ ReverseEOL} &\textbf{33.29} &\textbf{61.56} &\textbf{40.53} &\textbf{70.69} &\textbf{73.14} &\textbf{50.43}\hgain{4.18}\\
\bottomrule
\end{tabular}}
\caption{Average performance on MTEB across five embedding task categories. \texttt{LLaMA-2-7B}, \texttt{LLaMA-2-13B}, and \texttt{Mistral-7B} refer to training-free embedding methods built upon these decoder-only LLMs. The best results are highlighted in \textbf{bold}. See Appendix~\ref{appendix:mteb_results} for per-dataset results.}
\label{tab:mteb_compare}
\end{minipage}%
\hfill
\begin{minipage}[t]{0.43\textwidth}
\centering
\scalebox{0.83}{%
\small
\setlength{\tabcolsep}{6pt}
\begin{tabular}{lcc}
\toprule
\textbf{Embedding Combination} & \textbf{STS} & \textbf{MTEB} \\
\midrule
\multicolumn{3}{l}{\textit{Single embedding (reversed)}} \\
\textit{\textbf{R}}-PromptEOL & 71.30 & 44.48 \\
\textit{\textbf{R}}-CoT & 72.58 & 46.23 \\
\textit{\textbf{R}}-Knowledge & 73.24 & 46.39 \\
\midrule
\multicolumn{3}{l}{\textit{Single embedding (forward)}} \\
PromptEOL & 74.88 & 44.57 \\
CoT & 77.23 & 46.56 \\
Knowledge & 77.14 & 47.32 \\
\midrule
\multicolumn{3}{l}{\textit{Two forward embeddings}} \\
PromptEOL + CoT & 76.79\drop{0.44} & 46.39\drop{0.17} \\
PromptEOL + Knowledge & 77.56\gain{0.42} & 47.17\drop{0.15} \\
CoT + Knowledge & 78.42\gain{1.19} & 47.59\gain{0.27} \\
\midrule
\multicolumn{3}{l}{\textit{Forward + reversed (\textbf{Ours})}} \\
\rowhcolor PromptEOL + \textit{\textbf{R}}-PromptEOL & 78.69\hgain{3.81} & 49.91\hgain{5.34} \\
\rowhcolor CoT + \textit{\textbf{R}}-CoT & 79.79\hgain{2.56} & 50.95\hgain{4.39} \\
\rowhcolor Knowledge + \textit{\textbf{R}}-Knowledge & 79.85\hgain{2.71} & 51.25\hgain{3.93} \\
\rowhcolor PromptEOL + \textit{\textbf{R}}-CoT & 78.81\hgain{3.93} & 50.41\hgain{4.18} \\
\rowhcolor CoT + \textit{\textbf{R}}-PromptEOL & 79.76\hgain{2.53} & 50.67\hgain{4.11} \\
\bottomrule
\end{tabular}}
\caption{Performance comparison of different embedding combinations on STS and MTEB using LLaMA-2-7B. \textit{\textbf{R}}-prefix denotes the reversed variant of the corresponding prompt; CoT refers to Pretended CoT.}
\label{tab:combine_ablation}
\end{minipage}
\end{table*}

\section{Analysis and Ablation}
\subsection{Mechanism Analysis}
\label{sec:mechanism_analysis}
\paragraph{Effect of Embedding Combinations.} A natural question is \textit{whether combining two strong forward embeddings can achieve comparable performance gains}. As shown in Table~\ref{tab:combine_ablation}, we compare several embedding combinations on STS and MTEB. Simply ensembling two forward embeddings not only requires additional prompt engineering but also yields marginal or even negative gains: for instance, combining PromptEOL with Pretended CoT ($76.79/46.39$) underperforms Pretended CoT alone ($77.23/46.56$). Moreover, other ensemble-based approaches, such as CP (introducing an auxiliary prompt) and MetaEOL (aggregating eight carefully designed prompts), still fall significantly short of ReverseEOL (see Tables~\ref{tab:sts_compare} and \ref{tab:mteb_compare}).

In contrast, combining any forward embedding with its \emph{reversed} counterpart consistently delivers substantial improvements. For example, although PromptEOL alone achieves limited performance ($74.88/44.57$), "PromptEOL + \textit{\textbf{R}}-PromptEOL" ($78.69/49.91$) clearly outperforms the strongest forward-only combination "CoT + Knowledge" ($78.42/47.59$), while "Knowledge + \textit{\textbf{R}}-Knowledge" further surpasses this forward combination by $1.43$ and $3.66$ points on STS and MTEB. The same trend is observed in cross-prompt combinations such as "PromptEOL + \textit{\textbf{R}}-CoT" and "CoT + \textit{\textbf{R}}-PromptEOL", yielding gains ranging from $2.53$ to $4.18$ points. Overall, these results confirm that naively combining two forward embeddings is not an effective strategy; the success of ReverseEOL lies in complementing the original forward embedding with a reversed embedding.

\begin{figure*}[t]
    \centering
    \begin{subfigure}[b]{0.32\textwidth}
        \centering
        \includegraphics[width=\linewidth]{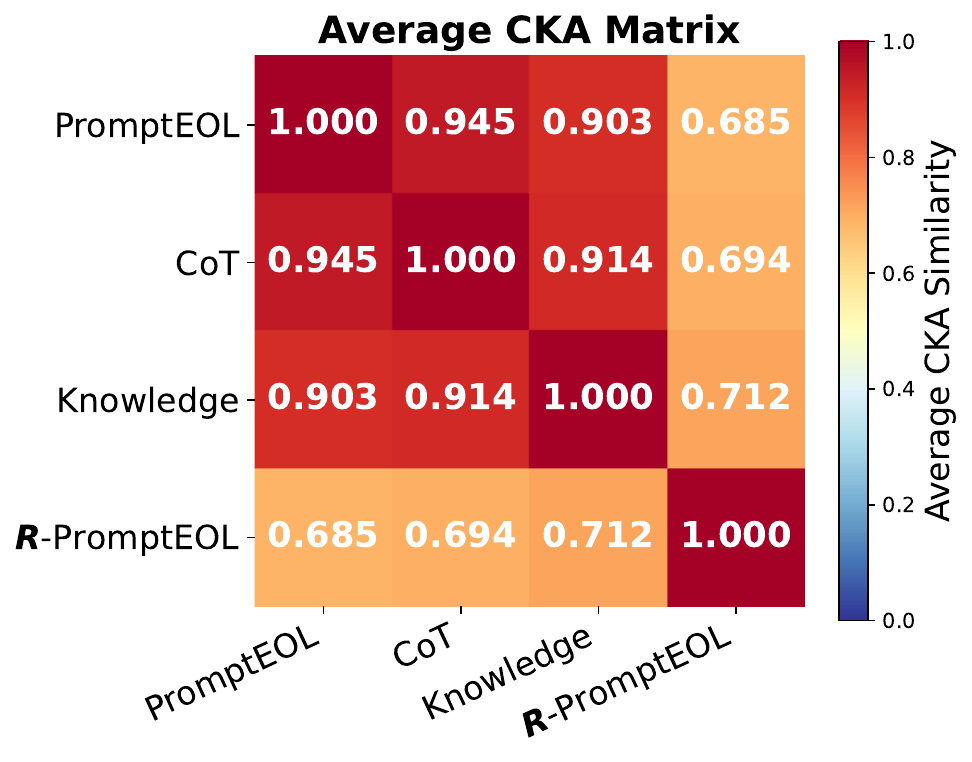}
        \caption{CKA similarity}
        \label{fig:cka}
    \end{subfigure}
    \hfill
    \begin{subfigure}[b]{0.32\textwidth}
        \centering
        \includegraphics[width=\linewidth]{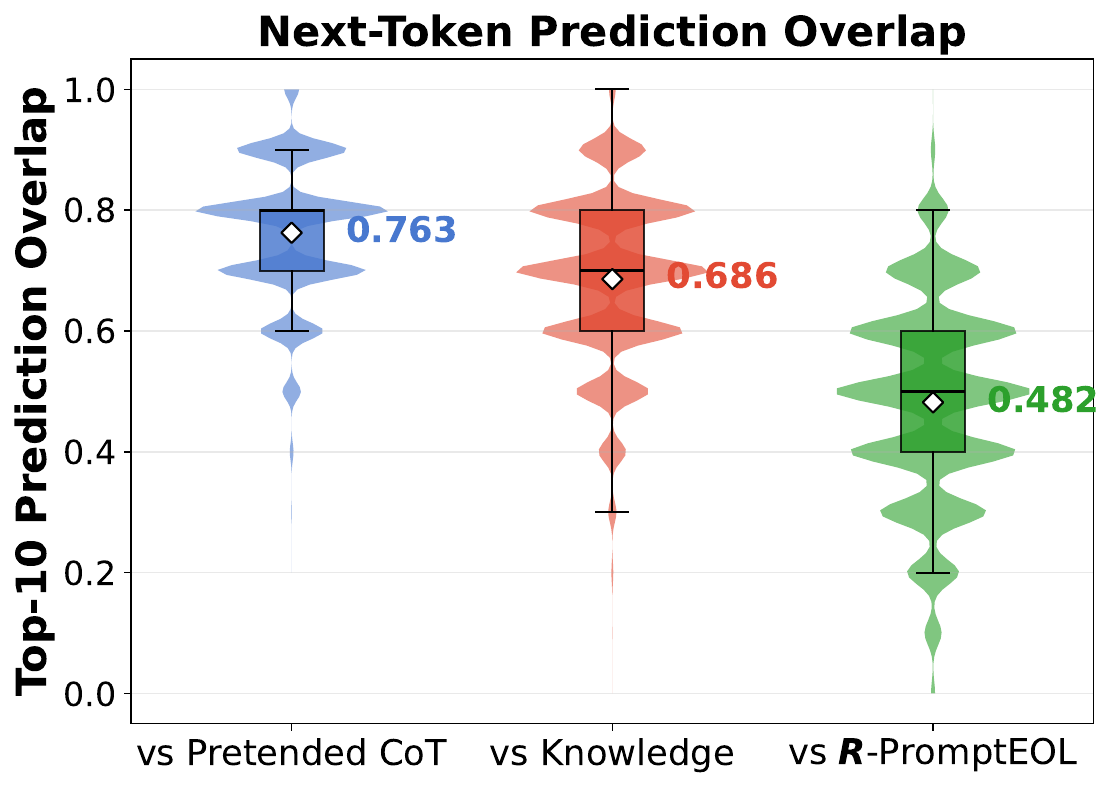}
        \caption{Next-token prediction overlap}
        \label{fig:token_overlap}
    \end{subfigure}
    \hfill
    \begin{subfigure}[b]{0.32\textwidth}
        \centering
        \includegraphics[width=\linewidth]{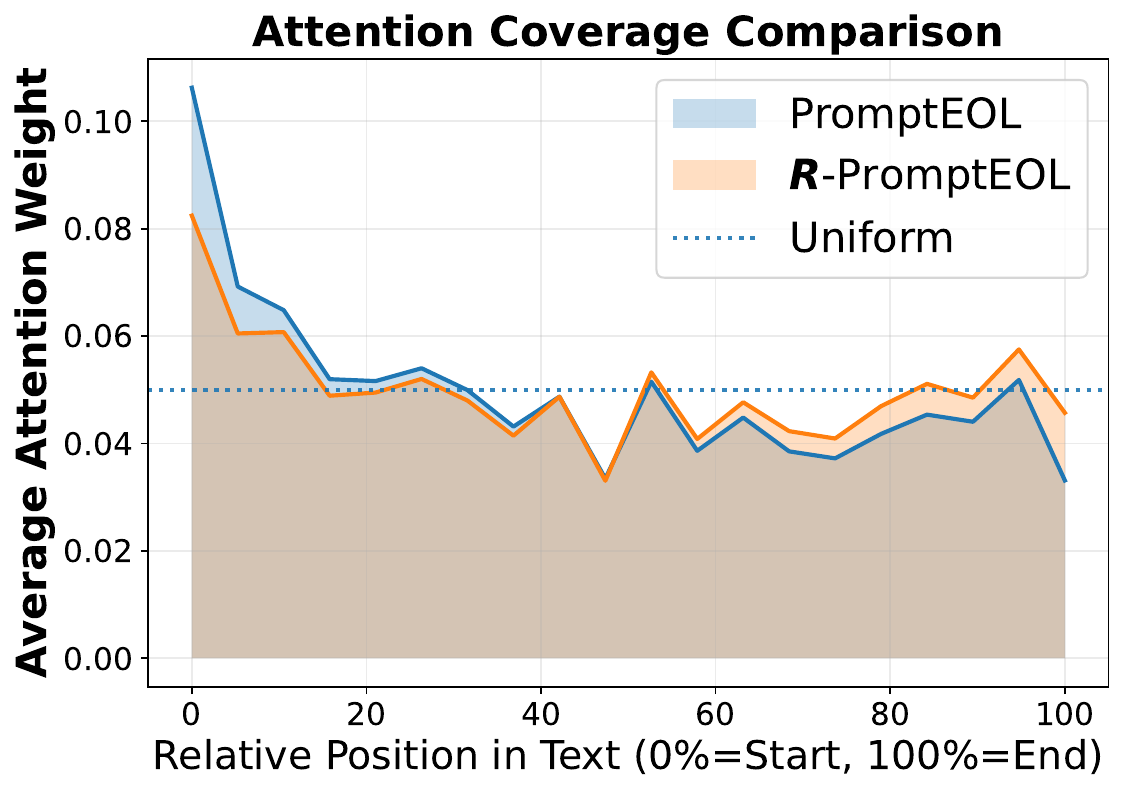}
        \caption{Attention distribution}
        \label{fig:attention}
    \end{subfigure}
    \caption{Analysis of forward and reversed embeddings using LLaMA-2-7B across MTEB datasets. (a) Pairwise CKA similarity. (b) Top-10 next-token prediction overlap between PromptEOL and other prompts. (c) Average attention distribution from the last token over relative positions of the input text. CoT refers to Pretended CoT.}
    \label{fig:analysis}
\end{figure*}

\paragraph{Reversed Embedding Encodes Complementary Information.} To understand the underlying mechanism of ReverseEOL, we compute Centered Kernel Alignment (CKA)~\cite{cka} between embeddings derived from different prompts and input orders. As shown in Figure~\ref{fig:cka}, PromptEOL, Pretended CoT, and Knowledge exhibit pairwise CKA scores above 0.9, suggesting that distinct prompts produce largely overlapping representations when applied to the same input order. In contrast, \textit{\textbf{R}}-PromptEOL shows markedly lower similarity to all forward embeddings, indicating that it encodes distinct contextual information unavailable in the original order. This explains the observation in Table~\ref{tab:combine_ablation}: combining different forward embeddings suffers from redundancy, whereas pairing a forward embedding with its reversed counterpart introduces complementary information, leading to significant improvements.

\paragraph{Next-Token Prediction Analysis.} Since training-free text embeddings are derived from the hidden state responsible for predicting the next token, we compare the top-10 next-token predictions produced by different prompt-based methods on MTEB datasets. As illustrated in Figure~\ref{fig:token_overlap}, PromptEOL shares 76.3\% and 68.6\% of its top-10 predictions with Pretended CoT and Knowledge, respectively. This confirms that no matter how carefully a prompt is designed, it is difficult to extract richer information from the same input. We hypothesize that the key to improving training-free embeddings lies in extracting as much semantic information as possible from the given text. To this end, simply reversing the input drives the model toward different predictive behaviors, reducing the overlap ratio to 48.2\%, with greater variance across samples. These results demonstrate that our reversal strategy is a remarkably effective solution for capturing diverse information.

\paragraph{Attention Pattern Analysis.} Figure~\ref{fig:attention} illustrates the average attention from the last token over the relative positions of the input text across MTEB datasets. The LLM exhibits nearly identical attention patterns for both forward and reversed texts, suggesting that it relies on similar positional inductive biases regardless of the input order. However, tokens at the same relative position differ across the two orders. More importantly, reversal allows each token to access a diverse preceding context under causal attention. Consequently, the last-token representation aggregates distinct semantic information under the two orders, which explains why the reversed embedding provides information complementary to the forward embedding.

\begin{table}[t]
\centering
\small
\setlength{\tabcolsep}{4pt}
\scalebox{0.90}{%
\begin{tabular}{lccc}
\toprule
\multirow{2}{*}{\textbf{Variant}} & \multicolumn{3}{c}{{Base Prompt}} \\
\cmidrule(lr){2-4}
 & \textbf{PromptEOL} & \textbf{CoT} & \textbf{Knowledge} \\
\midrule
\textit{w/o} ReverseEOL & 74.88 & 77.23 & 77.14 \\
\midrule
\rowhcolor ReverseEOL & \textbf{78.69} & \textbf{79.79} & \textbf{79.85} \\
\quad \textit{w/} Text-level Random & 76.74 & 78.25 & 78.01 \\
\quad \textit{w/} Token-level Reversal & 75.74 & 77.63 & 77.35 \\
\quad \textit{w/} Token-level Random & 75.89 & 77.54 & 77.54 \\
\bottomrule
\end{tabular}}
\caption{Comparison of ReverseEOL variants with different permutation strategies on STS tasks using LLaMA-2-7B. We evaluate three base prompts: PromptEOL, Pretended CoT, and Knowledge. Each variant applies a different permutation strategy to derive the additional embedding. Best results are in \textbf{bold}.}
\label{tab:permutation_ablation}
\end{table}

\subsection{Ablation Studies}
\label{ablation}
\paragraph{Effect of Different Permutation Strategies.}
By default, ReverseEOL applies text-level reversal (i.e., before tokenization) to obtain an additional embedding. To examine the importance of this permutation strategy, we further investigate three variants: (i) text-level random shuffling, (ii) token-level reversal (after tokenization), and (iii) token-level random shuffling. As shown in Table~\ref{tab:permutation_ablation}, both token-level variants substantially fall behind text-level permutations across all three training-free baselines. We attribute this to the fact that operating at the text level preserves semantic units (i.e., words) and provides more meaningful input for the LLM than operating on subword tokens. Moreover, at the text level, reversal clearly outperforms random shuffling, indicating that the structural complementarity it preserves, rather than mere stochastic permutation, is the key factor enabling the additional embedding to provide complementary information. Further analysis of text-level reversal and random shuffling is provided in Appendix~\ref{appendix:more_permuation}.

\begin{figure}[t]
\centering
\includegraphics[width=0.48\textwidth]{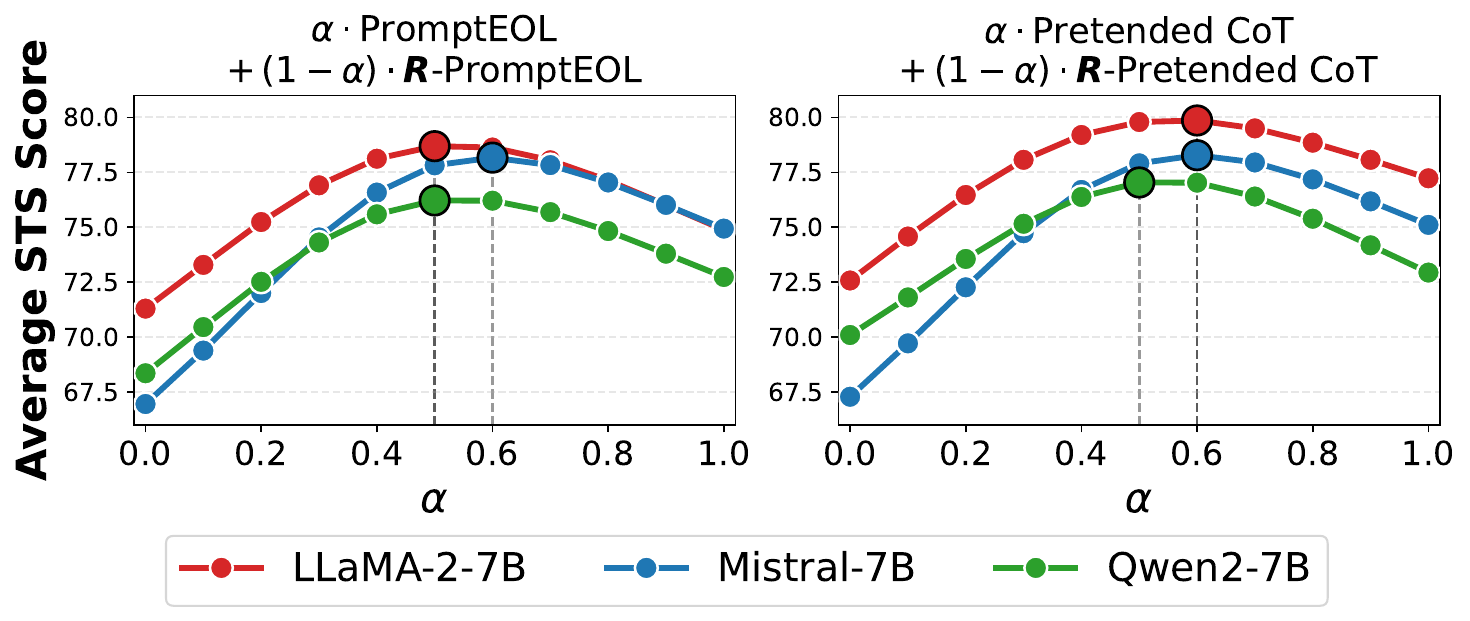}
\caption{Effect of the weighting coefficient $\alpha$ on STS tasks. \textit{\textbf{R}}-prefix denotes the reversed variant of the corresponding prompt.}
\label{fig:contribution}
\end{figure}

\paragraph{Contribution of Forward and Reversed Embeddings.} We analyze the contribution of the two embeddings within ReverseEOL by varying their weighting coefficient $\alpha$ in the final representation $\mathbf{e}^{\text{out}} = \alpha \cdot \mathbf{e}^{\text{fwd}} + (1-\alpha) \cdot \mathbf{e}^{\text{rev}}$. As shown in Figure~\ref{fig:contribution}, the optimal $\alpha$ consistently falls within $\{0.5, 0.6\}$, indicating that the forward and reversed embeddings contribute comparably to the final representation. To keep the method simple and parameter-free, we adopt averaging ($\alpha=0.5$) by default, which already yields substantial improvements over all baselines while avoiding the need for hyperparameter tuning across different methods or LLM backbones.

\begin{table*}[t]
\centering
\scalebox{0.79}{
\small
\setlength{\tabcolsep}{10pt}
\begin{tabular}{lccccccccc}
\toprule
\textbf{Method} & \textbf{Backbone} &\textbf{STS12} & \textbf{STS13} & \textbf{STS14} & \textbf{STS15} & \textbf{STS16} & \textbf{STS-B} & \textbf{SICK-R} & \textbf{Avg.}\\
\midrule
\multicolumn{10}{c}{\textbf{\texttt{Dense Models}}}\\\midrule
Pretended CoT & &57.89 &78.04 &65.02 &75.24 &76.89 &68.30 &64.81 &69.46 \\
\rowhcolor\quad \textit{w/ ReverseEOL} &\multirow{-2}{*}{\textbf{\texttt{Gemma-2-2B}}} &67.20 &81.30 &71.35 &80.23 &78.55 &72.43 &67.16 &74.03  \textcolor[rgb]{0.86,0.20,0.18}{\textbf{(+4.57)}}\\
\midrule
Pretended CoT & &64.27 &78.61 &69.93 &76.37 &79.28 &75.88 &69.04 &73.34  \\
\rowhcolor\quad \textit{w/ ReverseEOL} &\multirow{-2}{*}{\textbf{\texttt{LLaMA-2-13B}}} &71.22 &81.58 &74.13 &81.89 &82.19 &80.96 &72.23 &77.74  \textcolor[rgb]{0.86,0.20,0.18}{\textbf{(+4.40)}}\\
\midrule
Pretended CoT & &66.65 &82.60 &72.40 &79.36 &80.86 &77.09 &73.66 &76.09   \\
\rowhcolor\quad \textit{w/ ReverseEOL} &\multirow{-2}{*}{\textbf{\texttt{LLaMA-3-8B}}} &72.14 &83.62 &74.96 &82.27 &81.23 &79.38 &74.15 &78.25  \textcolor[rgb]{0.86,0.20,0.18}{\textbf{(+2.16)}}\\
\midrule
Pretended CoT & &61.64 &78.24 &70.14 &74.44 &76.63 &76.22 &73.30 &72.94   \\
\rowhcolor\quad \textit{w/ ReverseEOL} &\multirow{-2}{*}{\textbf{\texttt{Qwen2-7B}}} &68.43 &82.54 &74.18 &81.23 &79.14 &79.30 &74.45 &77.04  \textcolor[rgb]{0.86,0.20,0.18}{\textbf{(+4.10)}}\\
\midrule
Pretended CoT & &66.45  &82.04 &72.24 &77.93 &79.36 &76.66  &71.06 &75.11   \\
\rowhcolor\quad \textit{w/ ReverseEOL} &\multirow{-2}{*}{\textbf{\texttt{Mistral-7B}}} &71.04 &83.28 &74.91 &81.45 &81.51 &79.30 &73.86 &77.91  \textcolor[rgb]{0.86,0.20,0.18}{\textbf{(+2.80)}}\\
\midrule
\multicolumn{10}{c}{\textbf{\texttt{Mixture-of-Experts Models}}}\\\midrule
Pretended CoT & &64.09 &81.28 &69.93 &78.28 &76.70 &72.99 &68.06 &73.05 \\
\rowhcolor\quad \textit{w/ ReverseEOL} &\multirow{-2}{*}{\textbf{\texttt{OLMoE-1B-7B}}} &72.38 &82.71 &74.05 &81.63 &80.98 &78.04 &70.57 &77.19 \textcolor[rgb]{0.86,0.20,0.18}{\textbf{(+4.14)}}\\
\midrule
Pretended CoT & &59.24 &80.53 &68.55 &73.14 &73.95 &68.38 &68.55 &70.33  \\
\rowhcolor\quad \textit{w/ ReverseEOL} &\multirow{-2}{*}{\textbf{\texttt{MiniCPM-MoE-8x2B}}} &68.04 &81.56 &71.54 &77.73 &77.26 &76.16 &72.15 &74.92  \textcolor[rgb]{0.86,0.20,0.18}{\textbf{(+4.59)}}\\
\midrule
Pretended CoT & &65.81 &81.71 &69.83 &77.60 &78.60 &75.47 &71.91 &74.42   \\
\rowhcolor\quad \textit{w/ ReverseEOL} &\multirow{-2}{*}{\textbf{\texttt{Deepseek-MoE-16B}}} &70.11 &82.57 &72.36 &81.37 &80.81 &77.99 &74.59 &77.11  \textcolor[rgb]{0.86,0.20,0.18}{\textbf{(+2.69)}}\\
\midrule
Pretended CoT & &61.61 &81.26 &69.07 &75.34 &74.52 &75.10 &70.54 &72.49   \\
\rowhcolor\quad \textit{w/ ReverseEOL} &\multirow{-2}{*}{\textbf{\texttt{Qwen1.5-MoE-A2.7B}}} &66.52 &83.25 &73.66 &81.15 &77.43 &78.19 &72.36 &76.08  \textcolor[rgb]{0.86,0.20,0.18}{\textbf{(+3.59)}}\\
\midrule
Pretended CoT & &53.15 &70.97 &60.46 &73.35 &76.67 &73.66 &63.75 &67.43   \\
\rowhcolor\quad \textit{w/ ReverseEOL} &\multirow{-2}{*}{\textbf{\texttt{Qwen3-30B-A3B}}} &60.11 &75.86 &67.29 &74.95 &77.09 &75.64 &66.91 &71.12  \textcolor[rgb]{0.86,0.20,0.18}{\textbf{(+3.69)}}\\
\bottomrule
\end{tabular}}
\caption{Performance of ReverseEOL across 10 LLM backbones on STS tasks, covering both dense and Mixture-of-Experts (MoE) architectures. We adopt Pretended CoT as the base prompt for its simplicity and effectiveness.}
\label{tab:diiferent_backbone}
\end{table*}

\paragraph{Robustness across Different Backbones.} Since training-free text embeddings are directly elicited from LLMs without parameter updates, their applicability across different LLMs is critical in practice. However, most existing studies are limited to experiments on only a small number of LLMs. To thoroughly examine the generalizability of our method, we evaluate ReverseEOL on STS tasks across 10 LLMs spanning both dense and MoE architectures. As shown in Table~\ref{tab:diiferent_backbone}, ReverseEOL yields average improvements of $2.16$ to $4.57$ points on dense models and $2.69$ to $4.59$ points on MoE models, demonstrating strong generalization across diverse LLM backbones.

\begin{figure}[h]
\centering
\includegraphics[width=0.49\textwidth]{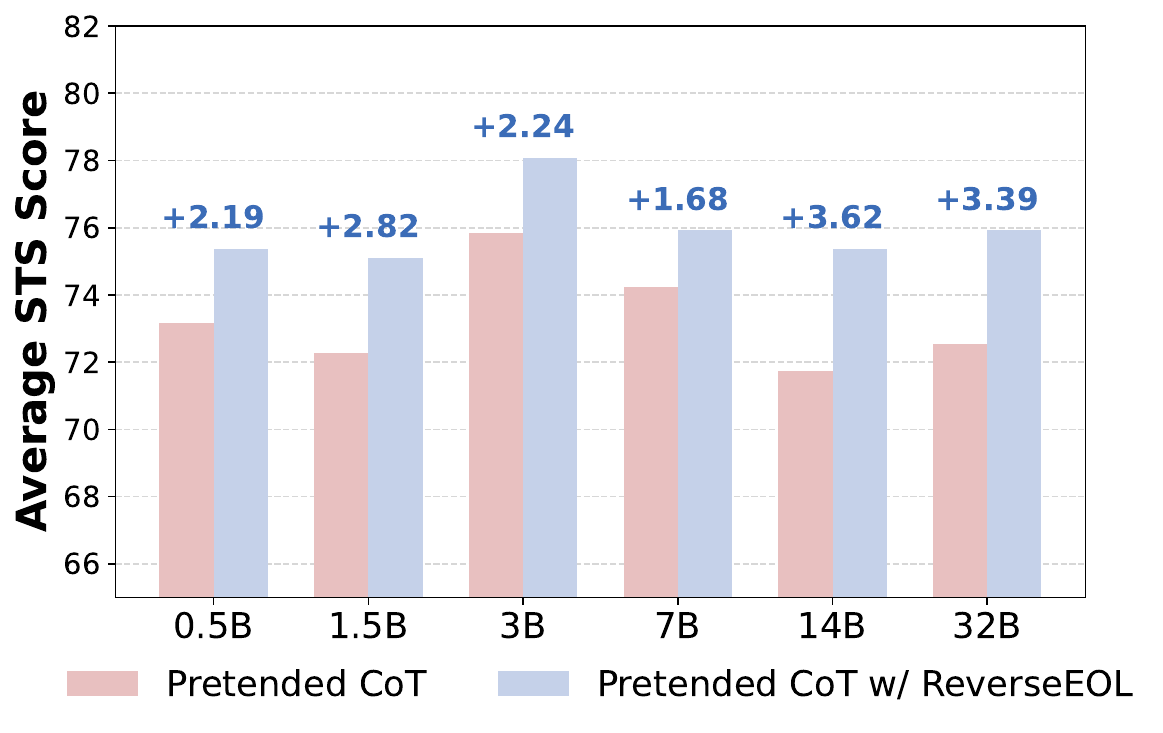}
\caption{Performance of ReverseEOL across different model scales on STS tasks, evaluated on the Qwen2.5 family with parameter sizes from 0.5B to 32B.}
\label{fig:scale}
\end{figure}

\paragraph{Robustness across Different Model Scales.} We further investigate whether the effectiveness of ReverseEOL is sensitive to model scale. Specifically, we evaluate ReverseEOL across the Qwen2.5 model family with parameter sizes spanning from 0.5B to 32B. As shown in Figure~\ref{fig:scale}, ReverseEOL consistently enhances performance across all model sizes, with improvements ranging from $1.68$ to $3.39$ points on STS tasks. Notably, the best performance is achieved by the 3B model rather than the largest one, suggesting that more parameters do not necessarily lead to higher-quality training-free embeddings. This observation is consistent with prior findings~\cite{cp,tp}.

\section{Conclusion}
In this paper, we introduced ReverseEOL, a simple yet effective method that enhances training-free text embeddings by augmenting the original forward embedding with a reversed counterpart obtained from the reversed input text. Reversing the input allows each token to access diverse preceding contexts under causal attention, enabling the resulting reversed embedding to effectively capture complementary semantic information for the forward one. As a result, combining the embeddings from two orders yields a higher-quality final representation. Extensive experiments on STS and MTEB across more than 10 LLM backbones demonstrated that ReverseEOL consistently improves training-free baselines by substantial margins, while requiring no complex prompt engineering or internal model interventions. We hope this work offers a new perspective for enhancing the representational capability of LLMs in a training-free manner.

\section*{Limitations}
While ReverseEOL significantly improves training-free text embeddings, it still has several limitations that motivate future research. First, since our work focuses on training-free embeddings, ReverseEOL does not match the performance of dedicated embedding models trained with contrastive learning on large-scale datasets. Bridging this gap while preserving the training-free paradigm remains an open challenge. Nevertheless, ReverseEOL offers a practical and attractive alternative in resource-constrained scenarios, and its model-agnostic nature allows it to benefit from continual advances in LLM capabilities. Second, our experiments primarily focus on English datasets, and extending the method to multilingual settings is a promising direction for future work.

\section*{Ethical Considerations}
Our study aims to enhance training-free text embeddings derived from LLMs, with potential benefits for downstream tasks such as retrieval-augmented generation and personalized recommendation. Since ReverseEOL builds upon frozen LLMs without parameter updates, any biases~\cite{shin-etal-2024-ask} or hallucinations~\cite{ravichander-etal-2025-halogen} inherent in the base model will inevitably propagate into the resulting embeddings. We recommend that users consider these risks when deploying the proposed method. In addition, all evaluation benchmarks used in this work are open-source resources commonly used by the community, which alleviates potential ethical risks related to data usage.

\bibliography{main}

\appendix
\section{Experiment Details}
\label{appendix:experimental_detail}
\subsection{MTEB Details}
\label{appendix:mteb_detail}
We utilize the English subset of the Massive Text Embedding Benchmark (MTEB)~\cite{mteb}, distributed under the \href{https://github.com/embeddings-benchmark/mteb/blob/main/LICENSE}{Apache License 2.0}, which comprises 32 datasets spanning 5 task categories: Retrieval (Retr.), Reranking (Rerank.), Clustering (Clust.), Pair Classification (PairClass.), and Classification (Class.). The corresponding evaluation metrics are nDCG@10, MAP, V-measure (V-meas.), average precision (AP), and accuracy (Acc.), respectively. The detailed composition of each task category is provided in Table~\ref{tab:mteb-composition}.

\subsection{STS Details}
\label{appendix:sts_detail}
The STS tasks we evaluate consist of seven datasets: STS 2012--2016~\cite{sts12,sts13,sts14,sts15,sts16}, STS-B~\cite{stsb}, and SICK-R~\cite{sickr}. Each dataset contains sentence pairs annotated with similarity scores ranging from 0 to 5. We report Spearman correlation between the predicted cosine similarity and the annotated similarity scores.

Notably, our STS evaluation pipeline is based on the official PromptEOL~\cite{prompteol} codebase, which is widely adopted by mainstream training-free text embedding methods~\cite{metaeol,cp,tp}, ensuring fair comparison and reproducibility. Therefore, we do not use the STS evaluation provided by the MTEB framework, as the two pipelines differ in implementation details and would lead to inconsistent results.

\begin{figure*}[t]
    \centering
    \begin{subfigure}[b]{0.32\textwidth}
        \centering
        \includegraphics[width=\linewidth]{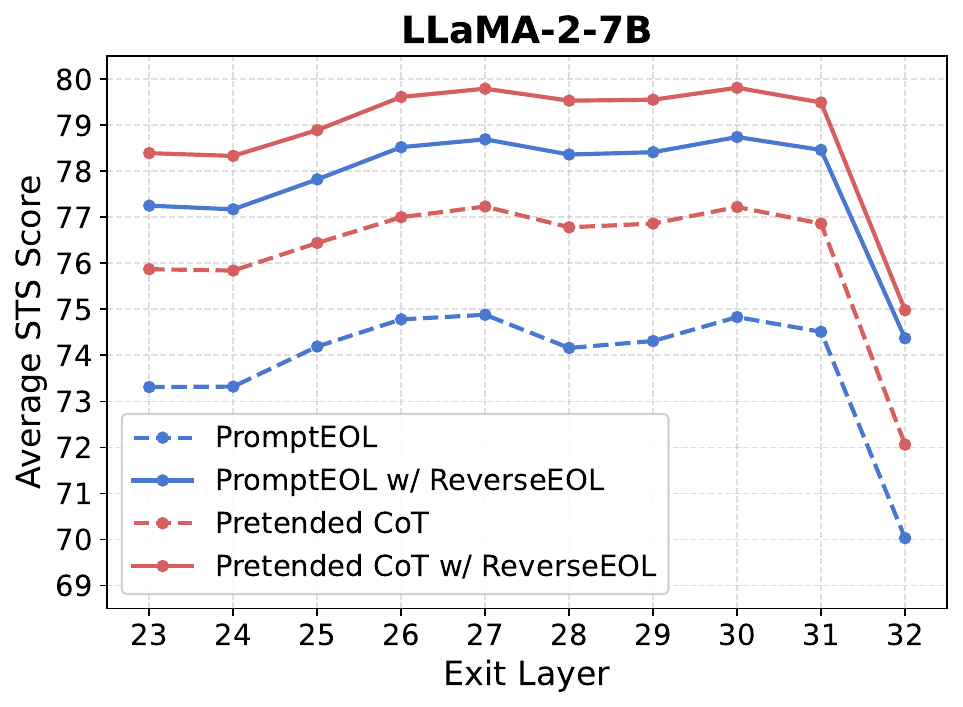}
        \caption{LLaMA-2-7B}
        \label{fig:layer_llama}
    \end{subfigure}
    \hfill
    \begin{subfigure}[b]{0.32\textwidth}
        \centering
        \includegraphics[width=\linewidth]{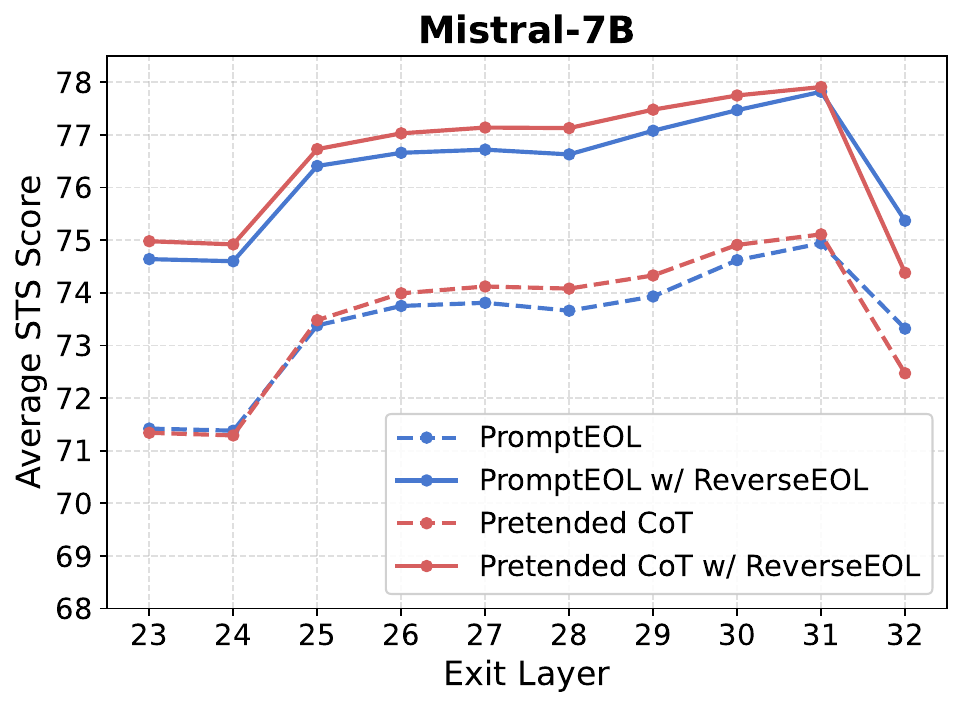}
        \caption{Mistral-7B}
        \label{fig:layer_mistral}
    \end{subfigure}
    \hfill
    \begin{subfigure}[b]{0.32\textwidth}
        \centering
        \includegraphics[width=\linewidth]{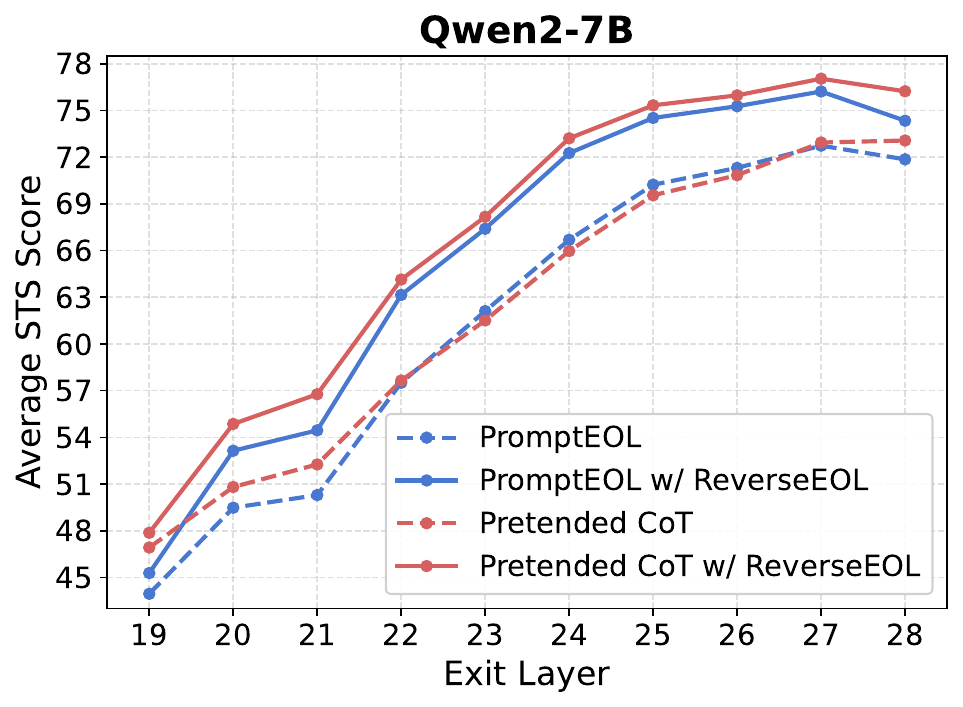}
        \caption{Qwen2-7B}
        \label{fig:layer_qwen}
    \end{subfigure}
    \caption{Effect of the exit layer $\ell$ for eliciting text embeddings on STS tasks. We evaluate three LLM backbones (LLaMA-2-7B, Mistral-7B, and Qwen2-7B) with PromptEOL and Pretended CoT as base prompts. ReverseEOL consistently improves over the corresponding training-free baselines at each layer.}
    \label{fig:output_layer}
\end{figure*}

\subsection{HuggingFace Models}
\label{appendix:llm_detail}
All LLMs employed in this work are obtained from the HuggingFace platform, as listed in Table~\ref{tab:hf_models}.

\subsection{Forward and Reversed Prompts}
\label{appendix:forward_reversed_prompt}
Our ReverseEOL can be readily integrated with existing training-free text embedding methods, as it does not require internal model interventions or additional complex prompt engineering. To derive the reversed embedding, our reversed prompt only requires a minimal modification to the original one: we insert the word ``reversed'' before ``sentence'' or ``text'' to indicate that the input has been reversed. Table~\ref{tab:appendix_prompts} lists various prompts with the one-word limitation and their corresponding reversed versions.

\section{Additional Results}
\subsection{Inference Speed Analysis}
We compare inference time across different methods on STS tasks, as shown in Table~\ref{tab:sts_time}. While ReverseEOL inevitably increases inference time ($1.96\times$) due to the additional forward pass required to compute the reversed embedding, it achieves the best overall performance. In contrast, prior methods that incur even higher computational cost fail to deliver comparable gains. For instance, MetaEOL introduces a significant $8.17\times$ overhead due to aggregating embeddings from eight prompts, yet its performance ($77.95$) still falls behind Pretended CoT \textit{w/} ReverseEOL ($79.79$). Overall, ReverseEOL offers a substantially better performance-efficiency trade-off than existing methods.

\begin{table}[h]
\centering
\scalebox{0.93}{%
\small
\setlength{\tabcolsep}{8pt}
\begin{tabular}{lcc}
\toprule
\textbf{Method} & \textbf{STS} & \textbf{Time} \\
\midrule
Pretended CoT & 77.23 & 1.00$\times$ \\
ECHO & 75.17 & 1.35$\times$ \\
Knowledge & 77.14 & 1.54$\times$ \\
Knowledge \texttt{w/} CP & 77.56 & 1.70$\times$ \\
MetaEOL & 77.95 & \textbf{8.17$\times$} \\
\rowhcolor Pretended CoT \texttt{w/} ReverseEOL (\textit{Ours}) & \textbf{79.79} & 1.96$\times$ \\
\bottomrule
\end{tabular}}
\caption{Comparison on STS tasks in terms of performance and inference time using LLaMA-2-7B.}
\label{tab:sts_time}
\end{table}

\subsection{Effect of the Exit Layer}
\label{appendix:exit_layer}
We analyze the effect of the exit layer $\ell$ on STS tasks using three LLM backbones 
(LLaMA-2-7B, Mistral-7B, and Qwen2-7B) and two prompt-based methods (PromptEOL 
and Pretended CoT). As shown in Figure~\ref{fig:output_layer}, eliciting training-free text embeddings from the last layer consistently yields suboptimal performance across all settings, which aligns with prior findings~\cite{liu-etal-2024-fantastic,jin-etal-2025-exploring} that intermediate layers tend to contain richer semantic information while the last layer is primarily optimized for next-token prediction. Notably, the penultimate layer is either optimal or very close to optimal across different methods and LLM backbones, and we therefore adopt it as the default exit layer for most LLMs in our experiments. An exception is LLaMA-2-7B, for which we follow prior work~\cite{tp,cp} and use the 27th layer to ensure fair comparison.

More importantly, ReverseEOL consistently delivers substantial performance improvements across all exit layers and both prompt-based methods, further confirming that the gains brought by the reversed embedding are robust to the choice of exit layer.

\subsection{Robustness to Prompt Design}
\label{appendix:prompt_robust}
To thoroughly verify the effectiveness of ReverseEOL across different prompts, in addition to the four prompts used in Table~\ref{tab:sts_compare}, we further include two prompts, \texttt{Prompt A} and \texttt{Prompt B}, derived from~\citet{prompta} and~\citet{promptb}. The full set of prompts along with their corresponding reversed variants is listed in Table~\ref{tab:appendix_prompts}.

It is worth noting that prior work often compares prompt-based methods without eliciting text embeddings from the same layer, leading to unfair comparisons. In contrast, we use the same exit layer $\ell$ for each baseline and its ReverseEOL counterpart, which is essential for faithfully demonstrating the effectiveness of our approach. As shown in Table~\ref{tab:prompt_roubust}, ReverseEOL consistently improves performance across all six prompts and four LLM families (Gemma-2-2B, LLaMA-2-7B, Qwen2.5-7B, and Mistral-7B), yielding average improvements of up to 8.09 points. These results confirm that ReverseEOL is robust to different one-word limitation prompts and serves as a general enhancement to a wide range of training-free embedding methods.

\subsection{Effect of Different Pooling Strategies}
We evaluate three pooling strategies, including last-token, mean, and weighted mean pooling~\cite{sgpt}, on STS and MTEB benchmarks. As shown in Tables~\ref{tab:pooling_ablation_sts} and~\ref{tab:pooling_ablation_mteb}, last-token pooling consistently achieves the best performance, which we attribute to the one-word limitation prompt that effectively encourages LLMs to condense the overall semantics into the last token. Based on this observation, we adopt last-token pooling to elicit text embeddings in all main experiments. Notably, ReverseEOL yields consistent improvements across all three pooling strategies on both benchmarks, confirming that the complementary information introduced by the reversed embedding is beneficial regardless of the choice of pooling strategy.

\begin{table}[h]
\centering
\small
\setlength{\tabcolsep}{8pt}
\scalebox{0.93}{
\begin{tabular}{lcc}
\toprule
\textbf{Method} & \textbf{STS} & \textbf{MTEB} \\
\midrule
\multicolumn{3}{c}{\textit{Random Shuffle only (\textit{\textbf{RS}})}} \\
\midrule
\textit{\textbf{RS}}-PromptEOL           & 65.70 & 43.27 \\
\textit{\textbf{RS}}-Pretended CoT       & 68.90 & 45.06 \\
\textit{\textbf{RS}}-Knowledge           & 68.16 & 45.13 \\
\midrule
\multicolumn{3}{c}{\textit{Reversed only (\textit{\textbf{R}})}} \\
\midrule
\textit{\textbf{R}}-PromptEOL            & 71.30 & 44.48 \\
\textit{\textbf{R}}-Pretended CoT        & 72.58 & 46.23 \\
\textit{\textbf{R}}-Knowledge            & 73.24 & 46.39 \\
\midrule
\multicolumn{3}{c}{\textit{Forward only}} \\
\midrule
PromptEOL                         & 74.88 & 44.57 \\
Pretended CoT                     & 77.23 & 46.56 \\
Knowledge                         & 77.14 & 47.32 \\
\midrule
\multicolumn{3}{c}{\textit{Forward + Random Shuffle}} \\
\midrule
PromptEOL + \textit{\textbf{RS}}-PromptEOL          & 76.74 & 48.21 \\
Pretended CoT + \textit{\textbf{RS}}-Pretended CoT  & 78.25 & 49.88 \\
Knowledge + \textit{\textbf{RS}}-Knowledge          & 78.01 & 49.98 \\
\midrule
\multicolumn{3}{c}{\textit{Forward + Reversed (Ours)}} \\
\midrule
\rowhcolor PromptEOL + \textit{\textbf{R}}-PromptEOL           & {78.69} & {49.91} \\
\rowhcolor Pretended CoT + \textit{\textbf{R}}-Pretended CoT   & {79.79} & {50.95} \\
\rowhcolor Knowledge + \textit{\textbf{R}}-Knowledge           & \textbf{79.85} & \textbf{51.25} \\
\bottomrule
\end{tabular}}
\caption{Comparison of reversal (\textit{\textbf{R}}) and random shuffle (\textit{\textbf{RS}}) on STS and MTEB tasks using LLaMA-2-7B.}
\label{tab:permutation_ablation_appendix}
\end{table}

\begin{figure}[h]
\centering
\includegraphics[width=0.47\textwidth]{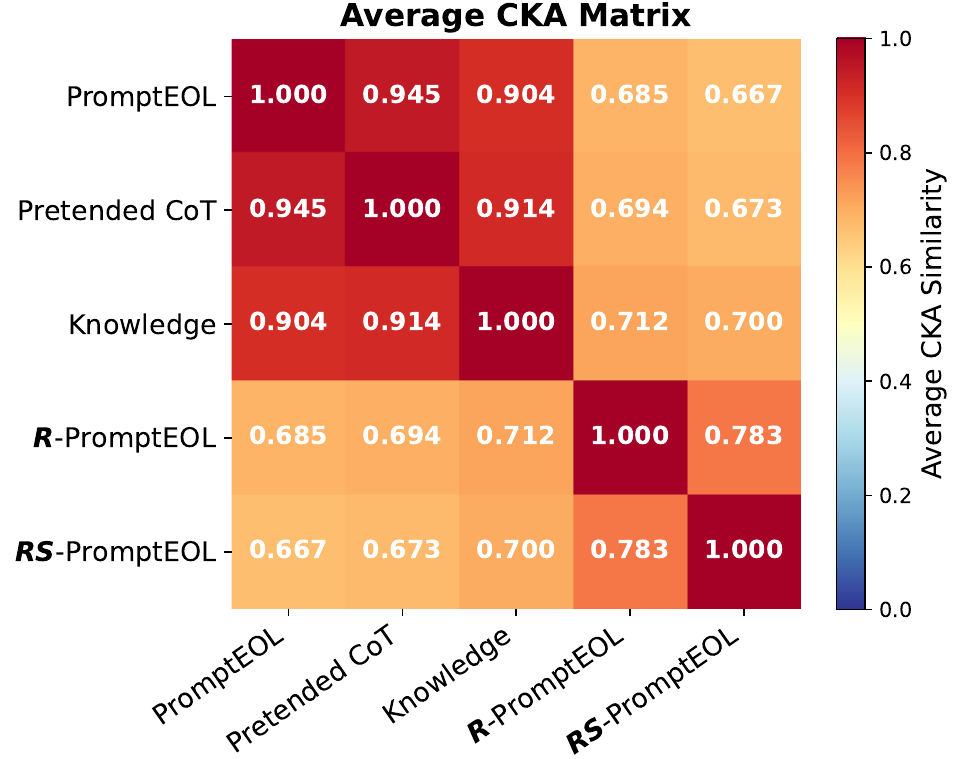}
\caption{Average CKA similarity matrix among five prompt-based methods across MTEB datasets using LLaMA-2-7B. \textit{\textbf{R}} and \textit{\textbf{RS}} denote the reversed and random-shuffle variants, respectively.}
\label{fig:cka_with_shuffle}
\end{figure}

\subsection{Reversal vs. Random Shuffling}
\label{appendix:more_permuation}
Building on the analysis in Section~\ref{ablation}, we further compare reversal \textit{\textbf{R}} and random shuffle \textit{\textbf{RS}} on the input text from both representational and performance perspectives. As shown in Figure~\ref{fig:cka_with_shuffle}, both permutation strategies produce embeddings with low CKA similarity to forward embeddings, ranging from $0.67$ to $0.71$, which is substantially lower than the similarity above $0.90$ observed among forward embeddings produced by different prompts. Consequently, as demonstrated in Table~\ref{tab:permutation_ablation_appendix}, combining the forward embedding with either variant improves performance on STS and MTEB over the forward-only baselines, confirming that both strategies provide semantic information that is inaccessible under the original forward order.

In particular, combining the forward embedding with the \textit{\textbf{R}}-embedding consistently outperforms combining with the \textit{\textbf{RS}}-embedding across all three base prompts and both benchmarks, despite \textit{\textbf{RS}}-embeddings exhibiting lower CKA to forward embeddings. This indicates that lower representational similarity does not necessarily lead to more useful complementary information. We attribute this gap to the fact that reversal preserves the sequential structure of the input, whereas random shuffling disrupts local dependencies and introduces incoherent contexts under causal attention. The relatively high CKA of $0.78$ between \textit{\textbf{R}} and \textit{\textbf{RS}} embeddings further highlights that the two strategies share similar semantic information, while the remaining gap reflects the more valuable complementary information uniquely provided by reversal. These results confirm the necessity and effectiveness of text reversal as the core mechanism of ReverseEOL.

\subsection{Detailed MTEB Results}
\label{appendix:mteb_results}
In this work, we evaluate ReverseEOL on 32 datasets from the English subset of MTEB, covering five task categories: Classification (7 datasets), Clustering (10 datasets), Retrieval (9 datasets), Pair Classification (3 datasets), and Reranking (3 datasets). The detailed per-dataset results of ReverseEOL and other training-free baselines are reported in Tables~\ref{tab:appendix_mteb_classification},~\ref{tab:appendix_mteb_clustering},~\ref{tab:appendix_mteb_retr},~\ref{tab:appendix_mteb_paircls}, and~\ref{tab:appendix_mteb_reranking}.

\begin{table*}[h]
    \centering
    \small
    \scalebox{1}{
    \begin{tabular}{ll}
    \toprule
    Category & Dataset \\ \toprule
    \multicolumn{1}{l}{\multirow{2}{*}{Retrieval (9)}}&  ArguAna, SciFact, NFCorpus, FiQA2018, SCIDOCS, TRECCOVID, Touche2020,\\
    & QuoraRetrieval, CQADupstack\\
    \midrule
    
    \multicolumn{1}{l}{\multirow{1}{*}{Reranking (3)}} & AskUbuntuDupQuestions, SciDocsRR, StackOverflowDupQuestions\\
    \midrule

    \multicolumn{1}{l}{\multirow{3}{*}{Clustering (10)}} & BiorxivClusteringS2S, BiorxivClusteringP2P, MedrxivClusteringS2S,\\
    & MedrxivClusteringP2P, RedditClustering, RedditClusteringP2P, ArxivClusteringS2S\\
    & StackExchangeClustering, StackExchangeClusteringP2P, TwentyNewsgroupsClustering\\ 
    \midrule

    \multicolumn{1}{l}{\multirow{1}{*}{Pair Classification (3)}} & SprintDuplicateQuestions, TwitterURLCorpus, TwitterSemEval2015\\ 
    \midrule

    \multicolumn{1}{l}{\multirow{3}{*}{Classification (7)}} 
    & Banking77Classification, EmotionClassification, MassiveIntentClassification\\
    & MassiveScenarioClassification, MTOPDomainClassification, \\
    &ToxicConversationsClassification, TweetSentimentExtractionClassification\\
    \midrule
    
    \multicolumn{1}{l}{\multirow{1}{*}{Overall}} & 32 datasets\\
    \bottomrule
    \end{tabular}}
    \caption{Composition of the MTEB benchmark.}
    \label{tab:mteb-composition}
\end{table*}

\begin{table*}[h]
\centering
\small
\setlength{\tabcolsep}{16pt}
\begin{tabular}{ll}
\toprule
\textbf{Model} & \textbf{HuggingFace ID} \\
\midrule
\multicolumn{2}{c}{\textbf{\texttt{Dense Models}}} \\
\midrule
LLaMA-2-7B~\cite{llama2} & \href{https://huggingface.co/meta-llama/Llama-2-7b}{meta-llama/Llama-2-7b} \\
LLaMA-2-13B~\cite{llama2} & \href{https://huggingface.co/meta-llama/Llama-2-13b}{meta-llama/Llama-2-13b} \\
LLaMA-3-8B~\cite{llama3} & \href{https://huggingface.co/meta-llama/Meta-Llama-3-8B}{meta-llama/Meta-Llama-3-8B} \\
Qwen2-7B~\cite{qwen2} & \href{https://huggingface.co/Qwen/Qwen2-7B}{Qwen/Qwen2-7B} \\
Qwen2.5 Family~\cite{qwen2.5} & \href{https://huggingface.co/collections/Qwen/qwen25}{Qwen2.5 Collection} \\
Mistral-7B~\cite{mistral} & \href{https://huggingface.co/mistralai/Mistral-7B-v0.1}{mistralai/Mistral-7B-v0.1} \\
Gemma-2-2B~\cite{gemma2} & \href{https://huggingface.co/google/gemma-2-2b}{google/gemma-2-2b} \\
\midrule
\multicolumn{2}{c}{\textbf{\texttt{Mixture-of-Experts Models}}} \\
\midrule
OLMoE-1B-7B~\cite{olmoe} & \href{https://huggingface.co/allenai/OLMoE-1B-7B-0924}{allenai/OLMoE-1B-7B-0924} \\
MiniCPM-MoE-8x2B~\cite{minicpm} & \href{https://huggingface.co/openbmb/MiniCPM-MoE-8x2B}{openbmb/MiniCPM-MoE-8x2B} \\
Deepseek-MoE-16B~\cite{deepseekmoe} & \href{https://huggingface.co/deepseek-ai/deepseek-moe-16b-chat}{deepseek-ai/deepseek-moe-16b-chat} \\
Qwen1.5-MoE-A2.7B~\cite{qwen1.5} & \href{https://huggingface.co/Qwen/Qwen1.5-MoE-A2.7B}{Qwen/Qwen1.5-MoE-A2.7B} \\
Qwen3-30B-A3B~\cite{qwen3} & \href{https://huggingface.co/Qwen/Qwen3-30B-A3B}{Qwen/Qwen3-30B-A3B} \\
\bottomrule
\end{tabular}
\caption{LLMs employed in this work. All models are obtained from the HuggingFace platform.}
\label{tab:hf_models}
\end{table*}

\begin{table*}[h]
\centering
\small
\begin{tabular}{lp{6cm}p{6cm}}
\toprule
\textbf{Method} & \textbf{Forward Prompt} & \textbf{Reversed Prompt} \\
\midrule
PromptEOL & \texttt{This sentence : "[TEXT]" means in one word:"} & \texttt{This reversed sentence : "[REV\_TEXT]" means in one word:"} \\
\midrule
Pretended CoT & \texttt{After thinking step by step , this sentence : "[TEXT]" means in one word:"} & \texttt{After thinking step by step , this reversed sentence : "[REV\_TEXT]" means in one word:"} \\
\midrule
Knowledge & \texttt{The essence of a sentence is often captured by its main subjects and actions, while descriptive terms provide additional but less central details. With this in mind , this sentence : "[TEXT]" means in one word:"} & \texttt{The essence of a reversed sentence is often captured by its main subjects and actions, while descriptive terms provide additional but less central details. With this in mind , this reversed sentence : "[REV\_TEXT]" means in one word:"} \\
\midrule
ECHO & \texttt{Sentence : "[TEXT]" This sentence : "[TEXT]" means in one word:"} & \texttt{Sentence : "[TEXT]" This reversed sentence : "[REV\_TEXT]" means in one word:"} \\
\midrule
Prompt A & \texttt{Summarize sentence : "[TEXT]" in one word:"} & \texttt{Summarize reversed sentence : "[REV\_TEXT]" in one word:"} \\
\midrule
Prompt B & \texttt{The representative word for sentence : "[TEXT]" is:"} & \texttt{The representative word for reversed sentence : "[REV\_TEXT]" is:"} \\
\bottomrule
\end{tabular}
\caption{Forward and reversed prompts with the one-word limitation used in our experiments. \texttt{[TEXT]} denotes the original input text, and \texttt{[REV\_TEXT]} refers to its reversed version.}
\label{tab:appendix_prompts}
\end{table*}

\begin{table*}[t]
\centering
\scalebox{0.85}{
\small
\setlength{\tabcolsep}{10pt}
\begin{tabular}{lccccccccc}
\toprule
\textbf{Method} & \textbf{Pooling} &\textbf{STS12} & \textbf{STS13} & \textbf{STS14} & \textbf{STS15} & \textbf{STS16} & \textbf{STS-B} & \textbf{SICK-R} & \textbf{Avg.}\\
\midrule
PromptEOL & & 64.36 & 82.05 & 72.21 & 78.88 & 77.35 & 76.80 & 72.52 & 74.88\\
\rowhcolor\quad \textit{w/ ReverseEOL} &\multirow{-2}{*}{Last-token} & 70.71 & 84.57 & 76.14 & 83.39 & 80.65 & 81.72 & 73.64 & \textbf{78.69} \\
\midrule
PromptEOL &  & 44.23 & 48.75 & 47.80 & 60.18 & 55.78 & 46.58 & 57.27 & 51.51\\
\rowhcolor\quad \textit{w/ ReverseEOL} &\multirow{-2}{*}{Mean} & 58.11 & 58.13 & 57.61 & 70.70 & 68.75 & 62.47 & 62.98 & \textbf{62.68} \\
\midrule
PromptEOL &  & 48.49 & 58.72 & 52.97 & 64.70 & 64.37 & 55.65 & 59.87 & 57.82\\
\rowhcolor\quad \textit{w/ ReverseEOL} &\multirow{-2}{*}{Weighted Mean} & 59.00 & 61.48 & 58.44 & 70.91 & 69.69 & 64.48 & 63.43 & \textbf{63.92} \\
\midrule
Pretended CoT &  & 67.76 & 83.87 & 74.73 & 79.94 & 80.85 & 79.86 & 73.59 & 77.23\\
\rowhcolor\quad \textit{w/ ReverseEOL} &\multirow{-2}{*}{Last-token} & 72.20 & 85.51 & 77.51 & 83.92 & 82.10 & 82.80 & 74.50 & \textbf{79.79} \\
\midrule
Pretended CoT &  & 42.79 & 47.91 & 46.49 & 58.70 & 55.59 & 46.21 & 56.43 & 50.59\\
\rowhcolor\quad \textit{w/ ReverseEOL} &\multirow{-2}{*}{Mean} & 58.50 & 57.54 & 57.34 & 70.49 & 68.68 & 62.73 & 62.93 & \textbf{62.60} \\
\midrule
Pretended CoT &  & 46.84 & 57.99 & 51.56 & 63.39 & 63.75 & 54.92 & 58.87 & 56.76\\
\rowhcolor\quad \textit{w/ ReverseEOL} &\multirow{-2}{*}{Weighted Mean} & 59.57 & 60.95 & 58.17 & 70.75 & 69.58 & 64.75 & 63.26 & \textbf{63.86} \\
\bottomrule
\end{tabular}}
\caption{Effect of pooling strategies on STS tasks using LLaMA-2-7B. Best results are highlighted in \textbf{bold}.}
\label{tab:pooling_ablation_sts}
\end{table*}

\begin{table*}[t]
\centering
\scalebox{0.85}{
\small
\setlength{\tabcolsep}{12pt}
\begin{tabular}{lccccccc}
\toprule
\textbf{Method} & \textbf{Pooling} & \textbf{Retr.} & \textbf{Rerank.} & \textbf{Clust.} & \textbf{PairClass.} & \textbf{Class.} & \textbf{Avg.}\\
\midrule
PromptEOL & &26.92 &57.94 &31.94 &62.36 &71.97 &44.57 \\
\rowhcolor\quad \textit{w/ ReverseEOL} &\multirow{-2}{*}{Last-token} &\textbf{34.15} &\textbf{59.75} &\textbf{38.64} &\textbf{69.67} &\textbf{73.61} &\textbf{49.91} \\
\midrule
PromptEOL & &20.96 &52.28 &31.93 &66.19 &63.29 &40.82  \\
\rowhcolor\quad \textit{w/ ReverseEOL} &\multirow{-2}{*}{Mean} &\textbf{23.80} &\textbf{54.06} &\textbf{33.70} &\textbf{71.38} &\textbf{66.85} &\textbf{43.64}  \\
\midrule
PromptEOL & &23.73 & 53.89 & 32.59 & 65.14 & 65.30 &42.30  \\
\rowhcolor\quad \textit{w/ ReverseEOL} &\multirow{-2}{*}{Weighted Mean} &\textbf{26.39} &\textbf{54.70} &\textbf{34.94} &\textbf{70.68} &\textbf{67.29} &\textbf{44.81} \\
\midrule
Pretended CoT & &30.03 &60.37 &33.73 &65.13 &72.25 &46.56 \\
\rowhcolor\quad \textit{w/ ReverseEOL} &\multirow{-2}{*}{Last-token} &\textbf{36.22} &\textbf{61.34} &\textbf{38.98} &\textbf{71.17} &\textbf{73.85} &\textbf{50.95} \\
\midrule
Pretended CoT & &20.52 &52.40 & 32.09 & 66.38 &63.41 &40.81\\
\rowhcolor\quad \textit{w/ ReverseEOL} &\multirow{-2}{*}{Mean} &\textbf{24.04} &\textbf{54.30} &\textbf{33.95} &\textbf{71.91} &\textbf{66.80} &\textbf{43.81}\\
\midrule
Pretended CoT & &23.65 &53.97 &32.83 &65.27 &65.34 &42.38 \\
\rowhcolor\quad \textit{w/ ReverseEOL} &\multirow{-2}{*}{Weighted Mean} &\textbf{26.51} &\textbf{55.07} &\textbf{35.21} &\textbf{71.32} &\textbf{67.19} &\textbf{45.01} \\
\bottomrule
\end{tabular}}
\caption{Effect of pooling strategies on MTEB using LLaMA-2-7B. Best results are highlighted in \textbf{bold}.}
\label{tab:pooling_ablation_mteb}
\end{table*}

\begin{table*}[t]
\centering
\scalebox{0.86}{
\small
\setlength{\tabcolsep}{12pt}
\begin{tabular}{lcccccccc}
\toprule
\textbf{Method} & \textbf{STS12} & \textbf{STS13} & \textbf{STS14} & \textbf{STS15} & \textbf{STS16} & \textbf{STS-B} & \textbf{SICK-R} & \textbf{Average}\\
\midrule
\multicolumn{9}{c}{\textbf{\texttt{Gemma-2-2B}}}\\\midrule
PromptEOL &56.63 &79.23 &67.03 &75.06 &76.10 &68.62 &65.18 &69.69 \\
\rowhcolor\quad \textit{w/ ReverseEOL} &66.46 &82.42 &72.22 &80.24 &78.89 &73.39 &68.33 &74.56\textcolor[rgb]{0.86,0.20,0.18}{\textbf{(+4.87)}} \\
\midrule
Pretended CoT &57.89 &78.04 &65.02 &75.24 &76.89 &68.30 &64.81 &69.46 \\
\rowhcolor\quad \textit{w/ ReverseEOL} &67.20 &81.30 &71.35 &80.23 &78.55 &72.43 &67.16 &74.03 \textcolor[rgb]{0.86,0.20,0.18}{\textbf{(+4.57)}}\\
\midrule
Knowledge &59.88 &81.90 &70.51 &78.49 &79.53 &75.53 &71.75 &73.94 \\
\rowhcolor\quad \textit{w/ ReverseEOL}&67.18 &83.50 &74.16 &81.27 &80.43 &76.48 &72.36 &76.48 \textcolor[rgb]{0.86,0.20,0.18}{\textbf{(+2.54)}}\\
\midrule
ECHO  &50.87 &77.33 &61.91 &70.62 &71.41 &61.62 &63.76 &65.36 \\
\rowhcolor\quad \textit{w/ ReverseEOL} &62.48 &80.74 &68.79 &77.21 &75.93 &68.66 &66.53 &71.48 \textcolor[rgb]{0.86,0.20,0.18}{\textbf{(+6.12)}}\\
\midrule
Prompt A &60.00 &75.89 &63.66 &75.09 &73.07 &72.98 &69.97 &70.09 \\
\rowhcolor\quad \textit{w/ ReverseEOL} &65.95 &77.83 &67.14 &78.06 &75.01 &73.03 &69.55 &72.37 \textcolor[rgb]{0.86,0.20,0.18}{\textbf{(+2.28)}}\\
\midrule
Prompt B &49.59 &75.05 &61.77 &69.38 &68.42 &63.76 &66.03 &64.86 \\
\rowhcolor\quad \textit{w/ ReverseEOL} &57.78 &76.60 &64.93 &74.91 &74.79 &68.99 &67.26 &69.32 \textcolor[rgb]{0.86,0.20,0.18}{\textbf{(+4.46)}}\\
\midrule
\multicolumn{9}{c}{\textbf{\texttt{LLaMA-2-7B}}}\\\midrule
PromptEOL &64.36 &82.05 &72.21 &78.88 &77.35 &76.80 &72.52 &74.88 \\
\rowhcolor\quad \textit{w/ ReverseEOL} &70.71 &84.57 &76.14 &83.39 &80.65 &81.72 &73.64 &78.69\textcolor[rgb]{0.86,0.20,0.18}{\textbf{(+3.81)}} \\
\midrule
Pretended CoT &67.76 &83.87 &74.73 &79.94 &80.85 &79.86 &73.59 &77.23 \\
\rowhcolor\quad \textit{w/ ReverseEOL} &72.20 &85.51 &77.51 &83.92 &82.10 &82.80 &74.50 &79.79 \textcolor[rgb]{0.86,0.20,0.18}{\textbf{(+2.56)}}\\
\midrule
Knowledge &65.60 &82.82 &74.48 &80.75 &80.13 &80.34 &75.89 &77.14 \\
\rowhcolor\quad \textit{w/ ReverseEOL}&71.50 &85.56 &77.39 &83.74 &82.30 &82.51 &75.98 &79.85 \textcolor[rgb]{0.86,0.20,0.18}{\textbf{(+2.71)}}\\
\midrule
ECHO  &67.38 &81.08 &71.20 &78.92 &77.61 &77.82 &72.21 &75.17 \\
\rowhcolor\quad \textit{w/ ReverseEOL} &72.24 &84.69 &75.15 &83.29 &81.10 &81.98 &74.93 &79.05 \textcolor[rgb]{0.86,0.20,0.18}{\textbf{(+3.88)}}\\
\midrule
Prompt A &62.26 &80.17 &71.38 &76.93 &77.00 &75.86 &71.05 &73.52 \\
\rowhcolor\quad \textit{w/ ReverseEOL} &67.82 &82.51 &74.72 &81.60 &79.48 &79.20 &72.96 &76.90 \textcolor[rgb]{0.86,0.20,0.18}{\textbf{(+3.38)}}\\
\midrule
Prompt B &56.31 &79.80 &67.32 &73.67 &71.90 &67.44 &64.32 &68.68 \\
\rowhcolor\quad \textit{w/ ReverseEOL} &65.16 &82.67 &72.43 &79.81 &77.34 &74.18 &67.13 &74.10 \textcolor[rgb]{0.86,0.20,0.18}{\textbf{(+5.42)}}\\
\midrule
\multicolumn{9}{c}{\textbf{\texttt{Qwen2.5-7B}}}\\\midrule
PromptEOL &66.56 &80.65 &71.74 &79.25 &77.59 &78.74 &72.93 &75.35 \\
\rowhcolor\quad \textit{w/ ReverseEOL} &69.61 &82.01 &73.80 &81.55 &79.11 &79.13 &74.15 &77.05\textcolor[rgb]{0.86,0.20,0.18}{\textbf{(+1.70)}} \\
\midrule
Pretended CoT &66.71 &78.59 &71.21 &77.12 &76.02 &77.20 &72.81 &74.24 \\
\rowhcolor\quad \textit{w/ ReverseEOL} &68.58 &80.05 &72.59 &79.93 &77.94 &77.26 &75.09 &75.92 \textcolor[rgb]{0.86,0.20,0.18}{\textbf{(+1.68)}}\\
\midrule
Knowledge &62.76 &77.25 &69.10 &77.64 &73.52 &74.88 &72.43 &72.51 \\
\rowhcolor\quad \textit{w/ ReverseEOL}&66.74 &79.61 &71.55 &80.01 &75.95 &75.96 &74.40 &74.89 \textcolor[rgb]{0.86,0.20,0.18}{\textbf{(+2.38)}}\\
\midrule
ECHO  &68.12 &80.13 &71.58 &79.30 &77.40 &77.20 &72.76 &75.21 \\
\rowhcolor\quad \textit{w/ ReverseEOL} &68.40 &80.69 &71.63 &80.84 &78.59 &77.78 &73.44 & 75.91\textcolor[rgb]{0.86,0.20,0.18}{\textbf{(+0.70)}}\\
\midrule
Prompt A &65.33 &80.23 &71.69 &79.61 &75.45 &76.75 &73.21 &74.61 \\
\rowhcolor\quad \textit{w/ ReverseEOL} &70.03 &82.53 &74.63 &82.16 &78.74 &79.31 &75.14 &77.51 \textcolor[rgb]{0.86,0.20,0.18}{\textbf{(+2.90)}}\\
\midrule
Prompt B &60.90 &77.38 &63.29 &73.20 &73.08 &69.79 &68.48 &69.45 \\
\rowhcolor\quad \textit{w/ ReverseEOL} &65.23 &78.89 &68.56 &78.30 &74.77 &73.16 &68.61 &72.50 \textcolor[rgb]{0.86,0.20,0.18}{\textbf{(+3.05)}}\\
\midrule
\multicolumn{9}{c}{\textbf{\texttt{Mistral-7B}}}\\\midrule
PromptEOL &65.73 &80.77 &72.05 &78.09 &79.00 &77.69 &71.22 &74.94 \\
\rowhcolor\quad \textit{w/ ReverseEOL} &70.53 &82.71 &74.66 &81.56 &81.84 &79.60 &73.81 &77.82\textcolor[rgb]{0.86,0.20,0.18}{\textbf{(+2.88)}} \\
\midrule
Pretended CoT &66.45 &82.04 &72.24 &77.93 &79.36 &76.66 &71.06 &75.11 \\
\rowhcolor\quad \textit{w/ ReverseEOL} &71.04 &83.28 &74.91 &81.45 &81.51 &79.30 &73.86 &77.91 \textcolor[rgb]{0.86,0.20,0.18}{\textbf{(+2.80)}}\\
\midrule
Knowledge &60.33 &81.52 &71.73 &77.53 &77.99 &74.09 &74.02 &73.89 \\
\rowhcolor\quad \textit{w/ ReverseEOL}&65.34 &83.76 &74.72 &80.33 &79.85 &75.14 &73.09 &76.03 \textcolor[rgb]{0.86,0.20,0.18}{\textbf{(+2.14)}}\\
\midrule
ECHO  &62.97 &79.65 &69.97 &77.11 &75.57 &76.26 &71.65 &73.31 \\
\rowhcolor\quad \textit{w/ ReverseEOL} &69.16 &82.58 &73.52 &81.76 &79.58 &78.83 &74.72 & 77.16\textcolor[rgb]{0.86,0.20,0.18}{\textbf{(+3.85)}}\\
\midrule
Prompt A &65.74 &79.97 &73.00 &78.65 &78.21 &78.19 &74.24 &75.43 \\
\rowhcolor\quad \textit{w/ ReverseEOL} &70.60 &81.23 &73.71 &81.17 &80.14 &79.42 &74.76 &77.29 \textcolor[rgb]{0.86,0.20,0.18}{\textbf{(+1.86)}}\\
\midrule
Prompt B &41.47 &75.65 &59.87 &64.44 &68.41 &48.67 &58.71 &59.60 \\
\rowhcolor\quad \textit{w/ ReverseEOL} &57.29 &78.07 &64.50 &73.10 &75.16 &62.36 &63.37 &67.69 \textcolor[rgb]{0.86,0.20,0.18}{\textbf{(+8.09)}}\\
\bottomrule
\end{tabular}
}
\caption{Results on STS tasks for ReverseEOL applied to various prompt-based methods, including PromptEOL, Pretended CoT, Knowledge, Prompt A, and Prompt B; see Table~\ref{tab:appendix_prompts} for details of these prompts and their reversed variants. \texttt{LLaMA-2-7B}, \texttt{LLaMA-2-13B}, and \texttt{Mistral-7B} denote training-free embedding methods built upon these decoder-only LLMs. The best results are highlighted in \textbf{bold}.}
\label{tab:prompt_roubust}
\end{table*}

\begin{table*}[t]
\centering
\scalebox{0.85}{%
\small
\setlength{\tabcolsep}{7pt}
\begin{tabular}{lcccccccc}
\toprule
\textbf{Method} & \textbf{Banking77.} & \textbf{Emotion.} & \textbf{MassiveIn.} & \textbf{MassiveSc.} & \textbf{MTOPDom.} & \textbf{ToxicCon.} &\textbf{TweetSent.} &\textbf{Avg.}\\
\midrule
\multicolumn{9}{c}{\textbf{\texttt{LLaMA-2-7B}}}\\\midrule
ECHO & 82.36 & 46.08 & 72.81 & 77.20 & 93.22 & 71.93 & 57.61 &71.60 \\
PromptEOL & 80.06 & 49.18 & 74.59 & 77.74 & 91.76 & 69.49 & 60.98 &71.97\\
\quad \textit{w/ CP} & 80.59 & 49.34 & 75.16 & 78.69 & 92.12 & 70.22 & 61.02 &72.45\\
\quad \textit{w/ TP}  & 81.68 & 48.20 & 75.74 & 79.29 & 93.51 & 70.69 & 60.47 &72.80\\
\rowhcolor\quad \textit{w/ ReverseEOL} & 81.35 & 53.53 & 74.75 & 78.50 & 93.77 & 71.50 & 61.84 &\textbf{73.61}\\
\midrule
\multicolumn{9}{c}{\textbf{\texttt{LLaMA-2-13B}}}\\\midrule
ECHO & 81.01 & 42.75 & 70.64 & 74.64 & 91.15 & 65.59 & 55.09 & 68.70 \\
PromptEOL & 76.55 & 49.17 & 74.14 & 75.92 & 90.32 & 68.70 & 61.27 &70.87\\
\quad \textit{w/ CP}  & 76.09 & 48.79 & 73.61 & 75.30 & 90.17 & 68.18 & 61.14 &70.47\\
\quad \textit{w/ TP}  & 77.68 & 49.73 & 75.12 & 77.37 & 91.43 & 69.76 & 61.71 &71.83\\
\rowhcolor\quad \textit{w/ ReverseEOL} & 79.68 & 51.70 & 74.52 & 77.69 & 92.58 & 71.00 & 62.54 &\textbf{72.81}\\
\midrule
\multicolumn{9}{c}{\textbf{\texttt{Mistral-7B}}}\\\midrule
ECHO & 79.11 & 38.44 & 73.66 & 79.24 & 91.49 & 65.55 & 52.61 &68.59 \\
PromptEOL & 78.31 & 48.22 & 74.82 & 77.25 & 92.13 & 68.38 & 61.72 &71.55\\
\quad \textit{w/ CP}  & 78.12 & 47.78 & 74.85 & 77.17 & 92.02 & 68.55 & 61.53 &71.43\\
\quad \textit{w/ TP}  & 78.23 & 47.69 & 75.35 & 77.83 & 92.30 & 68.48 & 62.11 &71.71\\
\rowhcolor\quad \textit{w/ ReverseEOL} & 80.66 & 51.72 & 74.98 & 78.41 & 93.32 & 70.26 & 62.63 &\textbf{73.14} \\
\bottomrule
\end{tabular}}
\caption{Detailed results on the classification tasks of MTEB, including Banking77Classification (Banking77.), EmotionClassification (Emotion.), MassiveIntentClassification (MIntent.), MassiveScenarioClassification (MScene.), MTOPDomainClassification (MTOPDom.), ToxicConversationsClassification (ToxicCon.), and TweetSentimentExtractionClassification (TweetSent.). Best results are highlighted in \textbf{bold}.}
\label{tab:appendix_mteb_classification}
\end{table*}

\begin{table*}[t]
\centering
\scalebox{0.79}{%
\small
\setlength{\tabcolsep}{6pt}
\begin{tabular}{lccccccccccc}
\toprule
\textbf{Method} & \textbf{ArxivS2S.} & \textbf{BioP2P.} & \textbf{BioS2S.} & \textbf{MedP2P.} & \textbf{MedS2S.} & \textbf{Reddit.} &\textbf{RedditP2P.} & \textbf{StackEx.} &\textbf{StackP2P.} &\textbf{Twenty.} &\textbf{Avg.}\\
\midrule
\multicolumn{12}{c}{\textbf{\texttt{LLaMA-2-7B}}}\\\midrule
ECHO & 42.73 & 30.29 & 33.00 & 28.85 & 27.85 & 37.44 & 51.08 & 64.13 & 32.58 & 32.65 & 38.06 \\
PromptEOL & 38.15 & 23.79 & 28.95 & 21.98 & 25.43 & 29.28 & 45.02 & 46.73 & 26.49 & 33.61 &31.94\\
\quad \textit{w/ CP} & 41.33 & 29.60 & 25.06 & 25.89 & 22.42 & 31.64 & 46.41 & 48.21 & 26.73 & 34.69 & 33.20\\
\quad \textit{w/ TP}  & 38.65 & 24.29 & 29.15 & 21.48 & 25.05 & 35.90 & 46.16 & 56.61 & 25.70 & 38.40 &34.14\\
\rowhcolor\quad \textit{w/ ReverseEOL} & 39.59 & 31.68 & 32.55 & 26.28 & 28.13 & 43.14 & 54.19 & 58.53 & 31.08 & 41.19 &\textbf{38.64}\\
\midrule
\multicolumn{12}{c}{\textbf{\texttt{LLaMA-2-13B}}}\\\midrule
ECHO & 33.73 & 29.43 & 24.97 & 25.19 & 24.29 & 25.87 & 46.69 & 54.48 & 32.11 & 23.08 & 31.98 \\
PromptEOL & 38.85 & 25.65 & 27.01 & 22.46 & 24.56 & 27.78 & 41.78 & 46.29 & 27.96 & 35.40 & 31.77\\
\quad \textit{w/ CP}  & 38.65 & 25.37 & 27.53 & 22.20 & 25.13 & 28.37 & 41.72 & 48.20 & 27.72 & 36.15 & 32.11\\
\quad \textit{w/ TP}  & 38.79 & 28.27 & 27.79 & 23.74 & 25.33 & 29.48 & 42.52 & 51.80 & 26.78 & 37.17 & 33.17\\
\rowhcolor\quad \textit{w/ ReverseEOL} & 39.65 & 29.61 & 30.21 & 26.07 & 26.40 & 37.67 & 50.39 & 59.06 & 30.55 & 39.91 & \textbf{36.95}\\
\midrule
\multicolumn{12}{c}{\textbf{\texttt{Mistral-7B}}}\\\midrule
ECHO & 39.08 & 31.02 & 24.01 & 26.71 & 24.85 & 29.20 & 51.34 & 63.17 & 29.55 & 23.96 & 34.29 \\
PromptEOL & 42.55 & 28.71 & 30.69 & 24.28 & 27.25 & 33.23 & 52.93 & 55.32 & 29.61 & 37.55 & 36.21\\
\quad \textit{w/ CP}  & 42.23 & 27.28 & 30.54 & 23.59 & 27.18 & 32.57 & 52.41 & 54.45 & 29.96 & 36.64 & 35.68\\
\quad \textit{w/ TP}  & 41.45 & 29.27 & 30.66 & 24.94 & 26.98 & 32.10 & 53.19 & 53.53 & 28.97 & 35.82 & 35.69\\
\rowhcolor\quad \textit{w/ ReverseEOL} & 42.27 & 33.66 & 33.75 & 27.51 & 29.05 & 42.79 & 57.97 & 62.69 & 32.71 & 42.88 & \textbf{40.53} \\
\bottomrule
\end{tabular}}
\caption{Detailed results on the clustering tasks of MTEB, including ArxivClusteringS2S (ArxivS2S.), BiorxivClusteringP2P (BioP2P.), BiorxivClusteringS2S (BioS2S.), MedrxivClusteringP2P (MedP2P.), MedrxivClusteringS2S (MedS2S.), RedditClustering (Reddit.), RedditClusteringP2P (RedditP2P.), StackExchangeClustering (StackEx.), StackExchangeClusteringP2P (StackP2P.), and TwentyNewsgroupsClustering (Twenty.). Best results are highlighted in \textbf{bold}.}
\label{tab:appendix_mteb_clustering}
\end{table*}

\begin{table*}[t]
\centering
\scalebox{0.81}{%
\small
\setlength{\tabcolsep}{8pt}
\begin{tabular}{lcccccccccc}
\toprule
\textbf{Method} & \textbf{ArguAna} & \textbf{SciFact} & \textbf{NFCorpus} & \textbf{FiQA.} & \textbf{SCIDOCS} & \textbf{TREC.} &\textbf{Touche.} & \textbf{Quora.} &\textbf{CQADup.} &\textbf{Avg.}\\
\midrule
\multicolumn{11}{c}{\textbf{\texttt{LLaMA-2-7B}}}\\\midrule
ECHO & 34.19 & 38.85 & 9.66 & 9.35 & 7.48 & 35.68 & 3.47 & 80.61 & 12.31 & 25.73 \\
PromptEOL & 29.16 & 46.64 & 22.90 & 13.27 & 13.45 & 27.38 & 1.63 & 71.26 & 16.57 & 26.92\\
\quad \textit{w/ CP} & 30.81 & 48.73 & 23.18 & 14.17 & 14.62 & 30.09 & 2.60 & 72.16 & 17.61 & 28.22\\
\quad \textit{w/ TP}  & 33.95 & 42.20 & 22.26 & 15.10 & 13.75 & 39.48 & 5.89 & 76.94 & 19.02 & 29.84\\
\rowhcolor\quad \textit{w/ ReverseEOL} & 43.59 & 54.53 & 26.11 & 14.95 & 15.13 & 43.05 & 7.23 & 80.45 & 22.34 & \textbf{34.15}\\
\midrule
\multicolumn{11}{c}{\textbf{\texttt{LLaMA-2-13B}}}\\\midrule
ECHO & 34.11 & 32.86 & 8.49 & 9.27 & 6.52 & 34.70 & 4.66 & 78.30 & 13.98 & 24.77 \\
PromptEOL & 31.19 & 44.48 & \textbf{21.94} & 10.42 & 14.55 & 21.70 & 2.42 & 68.60 & 15.64 & 25.66\\
\quad \textit{w/ CP}  & 29.96 & 42.40 & 21.32 & 9.97 & 14.18 & 22.75 & 1.91 & 68.49 & 15.04 & 25.11\\
\quad \textit{w/ TP}  & 34.46 & 45.11 & 18.28 & 10.34 & 13.78 & 24.58 & 2.67 & 72.51 & 14.78 & 26.28\\
\rowhcolor\quad \textit{w/ ReverseEOL} & 43.86 & 54.12 & 21.53 & 10.91 & 13.96 & 37.87 & 8.36 & 78.32 & 18.65 & \textbf{31.95}\\
\midrule
\multicolumn{11}{c}{\textbf{\texttt{Mistral-7B}}}\\\midrule
ECHO & 28.59 & 16.45 & 9.03 & 6.22 & 4.42 & 35.49 & 2.72 & 78.71 & 11.00 & 21.41 \\
PromptEOL & 24.41 & 45.52 & 21.03 & 15.68 & 15.22 & 34.19 & 4.56 & 72.42 & 19.48 & 28.06\\
\quad \textit{w/ CP}  & 22.05 & 43.22 & 20.56 & 14.70 & 14.87 & 33.37 & 4.29 & 71.97 & 19.26 & 27.14\\
\quad \textit{w/ TP}  & 27.05 & 48.84 & 20.55 & 14.04 & 14.57 & 33.06 & 4.87 & 72.71 & 18.17 & 28.21\\
\rowhcolor\quad \textit{w/ ReverseEOL} & 44.13 & 56.61 & 22.47 & 12.30 & 15.79 & 42.54 & 7.07 & 80.19 & 18.56 & \textbf{33.29}\\
\bottomrule
\end{tabular}}
\caption{Detailed results on the retrieval tasks of MTEB, including ArguAna, SciFact, NFCorpus, FiQA2018 (FiQA.), SCIDOCS, TRECCOVID (TREC.), Touche2020 (Touche.), QuoraRetrieval (Quora.), and CQADupstack (CQADup.). Best results are highlighted in \textbf{bold}.}
\label{tab:appendix_mteb_retr}
\end{table*}

\begin{table*}[t]
\centering
\scalebox{0.81}{%
\small
\setlength{\tabcolsep}{14pt}
\begin{tabular}{lcccc}
\toprule
\textbf{Method} & \textbf{SprintDuplicateQuestions} & \textbf{TwitterSemEval2015} & \textbf{TwitterURLCorpus} &\textbf{Avg.}\\
\midrule
\multicolumn{5}{c}{\textbf{\texttt{LLaMA-2-7B}}}\\\midrule
ECHO & 78.23 & 63.96 & 79.92 & \textbf{74.04} \\
PromptEOL & 39.81 & 67.58 & 79.70 & 62.36\\
\quad \textit{w/ CP} & 40.88 & 80.18 & 68.88 & 63.31\\
\quad \textit{w/ TP}  & 46.63 & 68.30 & 79.03 & 64.65\\
\rowhcolor\quad \textit{w/ ReverseEOL} & 60.87 & 67.32 & 80.81 & 69.67\\
\midrule
\multicolumn{5}{c}{\textbf{\texttt{LLaMA-2-13B}}}\\\midrule
ECHO & 76.02 & 61.48 & 78.60 & \textbf{72.03}\\
PromptEOL & 27.21 & 66.44 & 79.25 & 57.63\\
\quad \textit{w/ CP}  & 26.68 & 65.88 & 79.05 &57.21\\
\quad \textit{w/ TP}  & 33.19 & 68.74 & 80.54 & 60.82\\
\rowhcolor\quad \textit{w/ ReverseEOL} & 58.79 & 67.15 & 81.14 & 69.03\\
\midrule
\multicolumn{5}{c}{\textbf{\texttt{Mistral-7B}}}\\\midrule
ECHO & 73.73 & 61.09 & 76.84 & 70.55 \\
PromptEOL & 31.94 & 70.88 & 80.46 & 61.09\\
\quad \textit{w/ CP}  & 31.67 & 70.75 & 79.93 & 60.78\\
\quad \textit{w/ TP}  & 32.59 & 71.86 & 81.34 & 61.93\\
\rowhcolor\quad \textit{w/ ReverseEOL} & 61.25 & 69.87 & 80.94 & \textbf{70.69}\\
\bottomrule
\end{tabular}}
\caption{Detailed results on the pair classification tasks of MTEB, including SprintDuplicateQuestions, TwitterSemEval2015, and TwitterURLCorpus. Best results are highlighted in \textbf{bold}.}
\label{tab:appendix_mteb_paircls}
\end{table*}

\begin{table*}[t]
\centering
\scalebox{0.85}{%
\small
\setlength{\tabcolsep}{14pt}
\begin{tabular}{lcccc}
\toprule
\textbf{Method} & \textbf{AskUbuntuDupQuestions} & \textbf{SciDocsRR} & \textbf{StackOverflowDupQuestions} &\textbf{Avg.}\\
\midrule
\multicolumn{5}{c}{\textbf{\texttt{LLaMA-2-7B}}}\\\midrule
ECHO & 54.34 & 77.38 & 45.72 & 59.15 \\
PromptEOL & 55.25 & 75.91 & 42.65 & 57.94\\
\quad \textit{w/ CP} & 55.68 & 76.76 & 42.78 & 58.41\\
\quad \textit{w/ TP}  & 57.03 & 75.87 & 43.53 & 58.81\\
\rowhcolor\quad \textit{w/ ReverseEOL} & 56.30 & 78.31 & 44.63 & \textbf{59.75}\\
\midrule
\multicolumn{5}{c}{\textbf{\texttt{LLaMA-2-13B}}}\\\midrule
ECHO & 52.44 & 65.11 & 41.69 & 53.08\\
PromptEOL & 56.24 & 78.18 & 41.36 & 58.59\\
\quad \textit{w/ CP}  & 56.16 & 78.38 & 41.55 & 58.70\\
\quad \textit{w/ TP}  & 56.87 & 79.03 & 43.11 & 59.67\\
\rowhcolor\quad \textit{w/ ReverseEOL} & 56.60 & 79.68 & 43.92 & \textbf{60.06}\\
\midrule
\multicolumn{5}{c}{\textbf{\texttt{Mistral-7B}}}\\\midrule
ECHO & 51.99 & 63.94 & 40.12 & 52.02 \\
PromptEOL & 56.94 & 80.26 & 43.95 & 60.39\\
\quad \textit{w/ CP}  & 57.37 & 80.04 & 44.22 & 60.54\\
\quad \textit{w/ TP}  & 57.08 & 79.89 & 43.57 & 60.18\\
\rowhcolor\quad \textit{w/ ReverseEOL} & 57.49 & 81.04 & 46.15 & \textbf{61.56} \\
\bottomrule
\end{tabular}}
\caption{Detailed results on the reranking tasks of MTEB, including AskUbuntuDupQuestions, SciDocsRR, and StackOverflowDupQuestions. Best results are highlighted in \textbf{bold}.}
\label{tab:appendix_mteb_reranking}
\end{table*}

\end{document}